\documentclass[10pt,twocolumn,letterpaper]{article}

\usepackage[pagenumbers]{cvpr} %

\usepackage[accsupp]{axessibility}
\usepackage{graphicx}
\usepackage{amsmath}
\usepackage{amssymb}
\usepackage{booktabs}

\usepackage{comment}
\usepackage{enumitem}

\newcommand{\eqdef}{\overset{\mathrm{def}}{=\joinrel=}}

\usepackage[pagebackref,breaklinks,colorlinks]{hyperref}

\usepackage[capitalize]{cleveref}
\crefname{section}{Sec.}{Secs.}
\Crefname{section}{Section}{Sections}
\Crefname{table}{Table}{Tables}
\crefname{table}{Tab.}{Tabs.}

\def\confName{CVPR}
\def\confYear{2023}

\begin{document}

\title{EventNeRF: Neural Radiance Fields from a Single Colour Event Camera}

\author{Viktor Rudnev\textsuperscript{1, 2}$\quad$
Mohamed Elgharib\textsuperscript{1}$\quad$
Christian Theobalt\textsuperscript{1}$\quad$
Vladislav Golyanik\textsuperscript{1}\vspace{7pt}\\
\textsuperscript{1}Max Planck Institute for Informatics, SIC \ \ \ \ \ \ \ \ \textsuperscript{2}Saarland University, SIC
}

\maketitle

\begin{abstract}

Asynchronously operating event cameras find many applications due to their high dynamic range, vanishingly low motion blur, low latency and low data bandwidth. 
The field saw remarkable progress during the last few years, and existing event-based 3D reconstruction approaches recover sparse point clouds of the scene.
However, such sparsity is a limiting factor in many cases, especially in computer vision and graphics, that has not been addressed satisfactorily so far.
Accordingly, this paper proposes the first approach for 3D-consistent, dense and photorealistic novel view synthesis using just a single colour event stream as input.
At its core is a neural radiance field trained entirely in a self-supervised manner from events while preserving the original resolution of the colour event channels. 
Next, our ray sampling strategy is tailored to events and allows for data-efficient training.
At test, our method produces results in the RGB space at unprecedented quality.
We evaluate our method qualitatively and numerically on several challenging synthetic and real scenes and show that it produces significantly denser and more visually appealing renderings than the existing methods.
We also demonstrate robustness in challenging scenarios with fast motion and under low lighting conditions.
We release the newly recorded dataset and our source code to facilitate the research field, see \url{https://4dqv.mpi-inf.mpg.de/EventNeRF}. 

\end{abstract}

\section{Introduction}

\begin{figure}[t!]
    \centering
    \footnotesize
    \includegraphics[width=\linewidth]{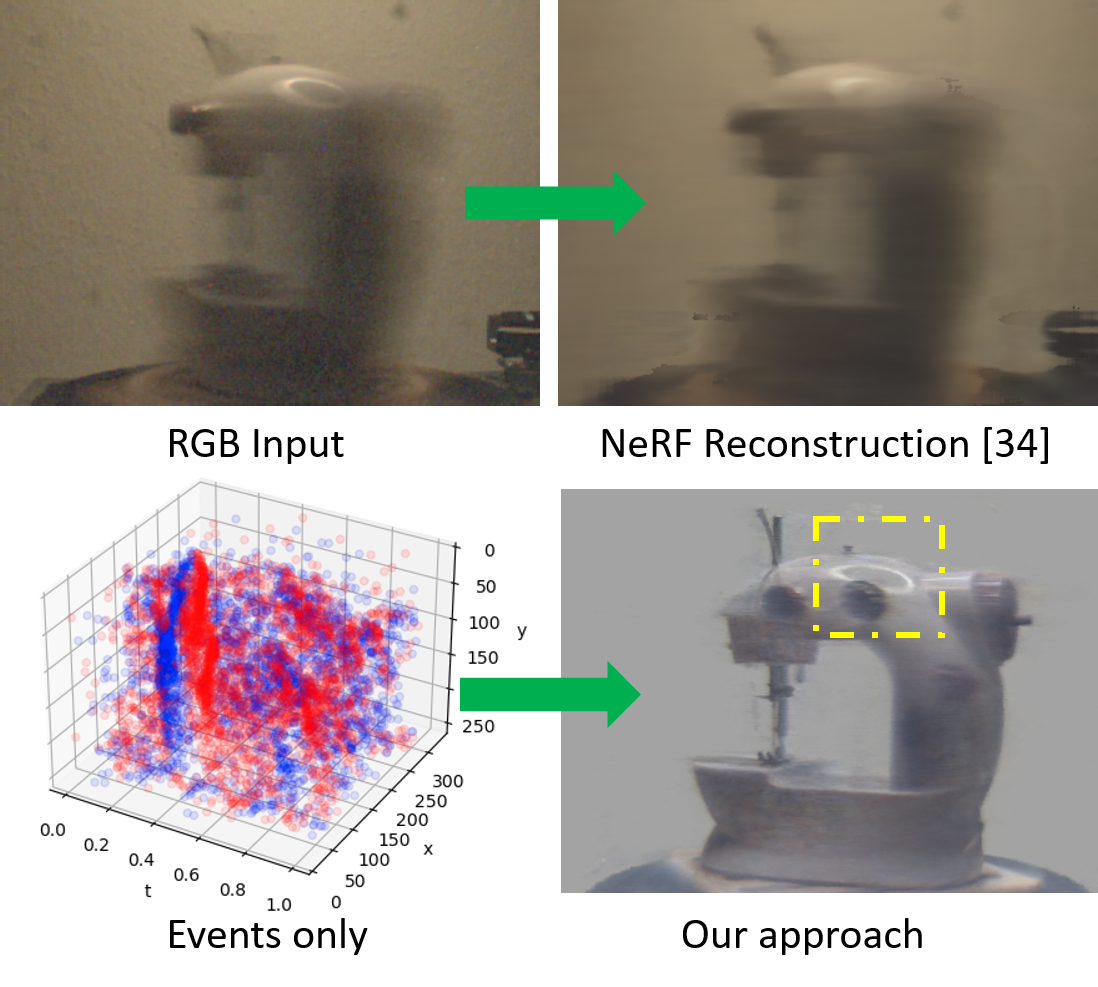}
    \vspace{-20pt}
    \caption{The original NeRF approach
    \cite{mildenhall2020nerf} (first row) produces strong artefacts when processing RGB images undergoing fast motion. Our approach (second row) operates purely on an event stream and thus produces significantly better and sharper results, capturing important details such as specularities (see yellow box). 
    }
    \label{fig:teaser}
\end{figure}

Event cameras record an asynchronous stream of per-pixel brightness changes.
This is in contrast to conventional RGB cameras that record the absolute intensity values at a pre-defined framerate.
Event cameras have several advantages over conventional cameras such as virtually no motion blur, higher dynamic range, ultra low power consumption and lower latency.
Because of these advantages, event cameras have been applied to many computer vision problems,
including optical-flow estimation~\cite{Gehrig22,Gehrig3dv2021}, video interpolation~\cite{pan2019bringing,Songnan20Learning,Tulyakov21CVPR,Wang_2021_ICCV}, deblurring~\cite{zhang2022unifying,Songnan20Learning}, 3D pose estimation~\cite{rudnev2021eventhands,Nehvi2021,EventCap2020,zou2021eventhpe}, geometry  reconstruction~\cite{Wang2022EvAC3D,Baudron20,Cui22,Uddin22}
and many more~\cite{Chen21,Wang_2021_ICCV,Stoffregen19,Zhou20,Wang20Restoration,Zhu19,Xiao22,Gallego22}.

Generating dense photorealistic rendering of a scene in a 3D-consistent manner is a long-standing problem in computer vision and computer graphics~\cite{Gortler96Lumigraph,Debevec96Modeling,Levoy96LFR,sitzmann2019srns,mildenhall2020nerf}.
There is currently, however, no event-based method for solving this problem.
The closest work in literature usually addresses the problem either from a SLAM~\cite{Zhou18eccv,Rebecq18,VidalSLAM,Kim16} or an odometry perspective~\cite{Zhou21tro,Zuo22,Hadviger21,Kueng16,Rebecq17EVO}.
Here, the aim is to estimate the position of a moving camera, and to reconstruct the 3D scene to some extend.
These methods usually rely on explicit feature matching in the event space and reconstruct sparse 3D models of environments from a single event stream.
As a result, novel views generated by existing methods are usually sparse, mostly containing edges and details causing events.
The same observation applies to methods using stereo event cameras, even though they aim to densify the 3D reconstructions~\cite{Zhou21tro,Zhou18eccv}.

The research question we are addressing in this work is, whether \textit{an event stream from an event camera moving around the scene is sufficient to reconstruct a dense volumetric 3D representation of a static scene}.
Here, we adopt the state-of-the-art 3D representation of Neural Radiance Fields (NeRF).
To this end, we present the first approach for inferring a NeRF volume from only a monocular colour event stream that enables 3D-consistent, dense and photorealistic novel view synthesis in the RGB space at test time.
Our method, which we name \textit{EventNeRF}, is designed for purely event-based supervision, during which we preserve the resolution of the individual RGB event channels.
We evaluate our technique for the task of novel view synthesis on a new dataset of synthetic and real event sequences.
This allows subjective and numerical evaluations against ground truth.
We evaluate our NeRF estimation in scenarios that would not be conceivable with a traditional RGB camera (\textit{e.g.,} high speed movements, motion blur or insufficient lighting), and show our method produce significantly better results (see Fig.~\ref{fig:teaser}).
We also show an application of extracting the depth map of an arbitrary viewpoint.
\begin{figure*}
    \centering
    \includegraphics[width=\textwidth,clip,trim={1cm 0 0 0}]{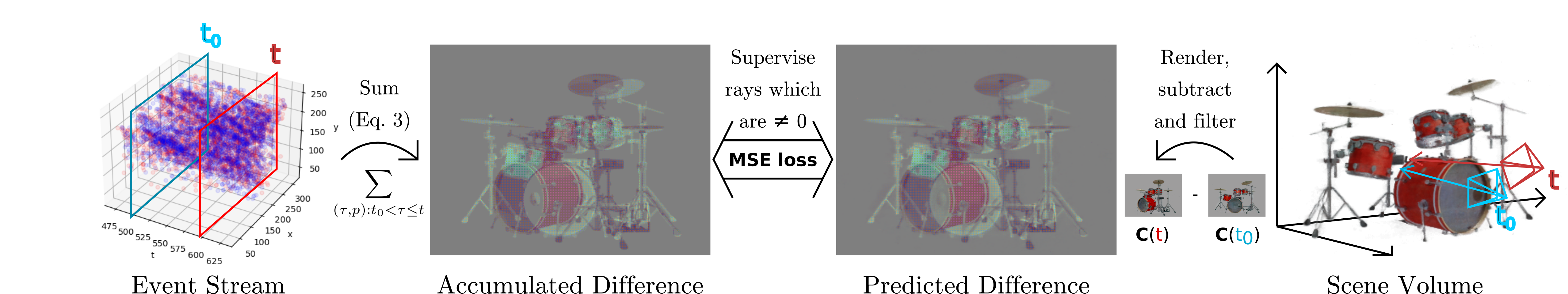}
    \caption{\textbf{Overview of the proposed method which learns a NeRF volume of a static object using just a moving event camera.} Our method connects the observed events $\{\mathbf{E}_i\}_{i=1}^{N}$ with the differences between rendered views (multiplied by Bayer colour filter) taken at two different time instants $t_0$ and $t$, via an event-based integral (see Sec.~\ref{sec:event_based_integral}).
    The projections are estimated via a fully differentiable  volumetric renderer (see Sec.~\ref{sec:NeRF}), thus allowing  learning the NeRF volume in a self-supervised manner.
    }
    \label{fig:method_overview}
\end{figure*}
To summarise, the primary \textbf{technical contributions} are as follows:
\begin{itemize}[leftmargin=15pt]
 \item[1)] EventNeRF, the first approach for inferring NeRF from a monocular colour event stream that enable novel view synthesis in the RGB space at test time.
 Our method is designed for event-based supervision, during which we preserve the resolution of the individual RGB event channels (while avoiding demosaicing; see Sec.~\ref{sec:colour}).
 \item[2)] A ray sampling strategy tailored to events that allows for data-efficient training. As a part of it, our negative sampling avoids artefacts in the background (Sec.~\ref{sec:ray_sampling}).
 \item[3)] An experimental evaluation protocol for the task of novel view  synthesis from event streams in the RGB space. We will release our code and dataset to  establish a benchmark for future work (Sec.~\ref{sec:experiments}).
\end{itemize}

\section{Related Work}

\noindent\textbf{Novel View Synthesis from Event Streams.}
The properties of event cameras (\textit{i.e.,} no motion blur, high dynamic range, low latency and ultra-low power consumption) motivated their usage in computer vision
~\cite{Tulyakov21CVPR,Wang20Restoration,zhang2022unifying,Gehrig22,Gehrig19ijcv}.
While many of the developed techniques address problems from a 2D-vision perspective~\cite{Gallego22}, event cameras have also been increasingly used for 3D-vision problems.
This includes methods designed for dedicated objects such as the human body~\cite{EventCap2020,zou2021eventhpe} and the human hands~\cite{rudnev2021eventhands,Nehvi2021}, as well as methods for general scenes~\cite{Baudron20,Wang2022EvAC3D,Cui22,Uddin22,VidalSLAM,Kim16,Rebecq18}.
EventCap~\cite{EventCap2020} is a method for the 3D reconstruction of a human body in motion.
Here, the tracked mesh can be used to render novel views of the performer.
However, the method requires grayscale frames and a 3D body scan as inputs. 

Thus, novel views of the actor do not contain new details observed in the input event streams (instead, the initial 3D scan is deforming).
The analysis-by-synthesis method of Nevhi~\textit{et  al.}~\cite{Nehvi2021} operates on events only and supports arbitrary non-rigid motions of the human hands.
Rudnev~\textit{et  al.}~\cite{rudnev2021eventhands} regresses sparse 3D hand keypoints and drives a MANO~\cite{MANO:SIGGRAPHASIA:2017} hand model from that.
Even though the final results of both Nevhi~\textit{et  al.}~\cite{Nehvi2021} or Rudnev~\textit{et  al.}~\cite{Nehvi2021} can be in form of a mesh, none of them is designed for the task of photorealistic novel view synthesis.

Methods for the 3D reconstruction of general scenes have recently seen remarkable progress.
They can be classified into ones that aim to estimate the underlying geometry only~\cite{Baudron20,Wang2022EvAC3D,Cui22,Uddin22} and others that also estimate the camera trajectory~\cite{VidalSLAM,Kim16,Rebecq18,Zhou21tro,Zuo22}. For the former, several methods have been proposed including ones that use a single event camera~\cite{Baudron20,Wang2022EvAC3D}, two event cameras~\cite{Uddin22} and other input data modalities such as LiDAR~\cite{Cui22}.
In the latter, methods follow either a SLAM~\cite{Zhou18eccv,Rebecq18,VidalSLAM,Kim16} or an odometry based formulation~\cite{Zhou21tro,Zuo22,Hadviger21,Kueng16,Rebecq17EVO} and target large-scale scenes. 
In addition to reconstructing the 3D scene to some extent, these methods also estimate the position of a moving camera scanning the scene.
To this end, some methods use as input a single event camera~\cite{Rebecq18,Kim16,Rebecq17EVO}, a pair of event cameras~\cite{Zhou18eccv,Zhou21tro,Hadviger21} and some use a mix of input data modalities such as depth~\cite{Zuo22}, intensity images~\cite{Kueng16} and IMUs~\cite{VidalSLAM}.
While SLAM aims for denser 3D reconstruction of the underlying scene than odometry methods, both  approach classes produce only sparse reconstructions (usually edges and corners).
This is far from allowing 3D-consistent dense and photorealistic rendering of the examined scene.
\textit{In stark contrast to these works, we learn---for the first time---an implicit 3D scene representation from event streams that enables dense photorealistic novel view synthesis in the RGB space.}

\vspace{4pt}\noindent\textbf{3D Scene Representation Learning}
is a long-standing problem in computer vision and computer graphics~\cite{Gortler96Lumigraph,Debevec96Modeling,Mildenhall19,sitzmann2019srns,Lombardi:2019,mildenhall2020nerf}.
Some of the existing methods produce meshes~\cite{Thies18Face2Face,DeepBlending2018,kair2022sft} and multi-plane images (MPIs)~\cite{Mildenhall19,zhou2018stereo,li2020crowdsampling}.
Such representations can be learnt from 2D images~\cite{Thies19DNR,zhou2018stereo,li2020crowdsampling},
but they suffer from several limitations, including the inability to capture fine geometrical details (meshes) and being mostly bounded to 2.5D novel view synthesis (MPIs).
In contrast, the recently introduced implicit coordinate-based representations learn to map the 3D spatial coordinates of a static scene to a continuous function of the local scene properties~\cite{Mescheder2019,Park2019,sitzmann2019srns,mildenhall2020nerf,Lombardi:2019}.
Here, Neural Radiance Fields, or NeRF~\cite{mildenhall2020nerf}, models a scene through MLP as a continuous function of the scene radiance and volume density learnt from a set of 2D RGB images.
At test time, NeRF accepts arbitrary 3D camera pose and viewing direction to render novel views.
Due to its simplicity and high accuracy of the generated novel views,
multiple follow-up works adopted NeRF for many applications such as reenactment~\cite{Gafni_2021_CVPR,liu2021neural,zheng2022imavatar}, scene appearance and light editing~\cite{martinbrualla2020nerfw,boss2021nerd,nerv2021}, 3D shape modelling~\cite{hong2021headnerf,corona2022lisa,chanmonteiro2020piGAN,Schwarz20} and  others~\cite{yang2021objectnerf,Niemeyer2020GIRAFFE,Tewari2022NeuRendSTAR}.
All these methods, however, assume input data captured by a traditional RGB camera.

\section{Method}

Our aim is to learn a neural 3D scene representation $\mathbf{C}(\mathbf{o},\mathbf{d})$ for a static scene from just a colour event stream $\{\mathbf{E}_i\}_{i=1}^{N}$ (no RGB images are involved).
The model $\mathbf{C}(\mathbf{o},\mathbf{d})$ is a function of camera position $\mathbf{o} \in \mathbb{R}^3$ and its viewing direction $\mathbf{d} \in \mathbb{R}^3, \|\mathbf{d}\|=1$, thus allowing novel view synthesis during test in the RGB space.
To learn the model, we assume the camera extrinsics $\mathbf{P}_j=[\mathbf{R}_j|\mathbf{t}_j] \in \mathbb{R}^{3\times4}$
and intrinsics $\textbf{K}_{j} \in \mathbb{R}^{3\times3}$ are known for each event window.
We also assume a constant colour background with a known value $\mathbf{\alpha} \in \mathbb{R}^{3}$ to rectify the reconstructed image brightness and colour balance.
Its known value is not a strict requirement as the equivalent brightness and colour balance corrections can be done after the rendering.
Our model is learned in a self-supervised manner, by comparing the approximate difference between views computed by summing the observed events polarities $\mathbf{E}(t_0, t)$ against the difference between the predicted views $\mathbf{L}(t)-\mathbf{L}(t_0)$ (Secs.~\ref{sec:volume_rendering}, \ref{sec:colour}).
To this end, %
we use NeRF \cite{mildenhall2020nerf} as our 3D scene representation, where each point in 3D-space  $\mathbf{x}$ is represented via a volume density
$\sigma(\mathbf{x})$ and a radiance $\mathbf{c}(\mathbf{x})$ (Sec.~\ref{sec:NeRF}).
We then use volumetric rendering~\cite{mildenhall2020nerf} to produce 2D projections of the learned
$\mathbf{C}(\mathbf{o},\mathbf{d})$, and establish a connection with the observed events via an event-based integral (Sec.~\ref{sec:event_based_integral}).
Since the volumetric rendering is fully differentiable, our model $\mathbf{C}(\mathbf{o},\mathbf{d})$ can be learned in a completely self-supervised manner using events only.
Our model is designed for rendering of novel view in the RGB space at test time; see Fig.~\ref{fig:method_overview}.
Lastly, we apply regularisation techniques and introduce an event-based ray sampling strategy that allows efficient learning of the model (Secs.~\ref{sec:distortion_reg}-\ref{sec:ray_sampling}).

\subsection{NeRF and Volumetric Rendering}
\label{sec:NeRF}
NeRF~\cite{mildenhall2020nerf} learns a volumetric model of a static scene from multiple 2D RGB inputs captured at different camera viewpoints.
It uses an MLP to map each point in 3D space into a continuous field of volume density and radiance.
To render a 2D image, a ray is shot from a camera location along a specific direction, and observations are projected into 2D via volumetric rendering~\cite{Kajiya85,Max95}.

Given a ray $\mathbf{r}(q)=\mathbf{o}+q\mathbf{d}$, where $\mathbf{o} \in \mathbb{R}^3$ is the camera origin and $\mathbf{d} \in \mathbb{R}^3, \|\mathbf{d}\|=1$ is the ray direction, we want to render its colour $\mathbf{C}(\mathbf{r}) \in \mathbb{R}^3$.
Random $S$ points $\{\mathbf{x}_i\}_{i=1}^S, \mathbf{x}_i \in \mathbb{R}^3$ on the ray are sampled with corresponding depths $\{t_i\}_{i=1}^S, t_i \in [t_n, t_f]$. Here, $t_n$ and $t_f$ are the near and the far camera plane depths.
The points $\{\mathbf{x}_i\}_{i=1}^S$ are encoded through positional encoding, akin to~\cite{sitzmann2019siren}:
We feed the encoded representation $\gamma(x)$
into an MLP $f_1(\cdot)$ to obtain the point densities $\{\sigma_i\}_{i=1}^S$ and feature vectors $\{\mathbf{u}_i\}_{i=1}^S$ via
$(\sigma_i, \mathbf{u}_i)=f_1(\gamma(\mathbf{x}_i))$.
We then combine the result with the view direction~$\mathbf{d}$ into the final colours of the ray samples $\{\mathbf{c}_i\}_{i=1}^S$ using MLP $f_2(\cdot)$ as
$\mathbf{c}_i = f_2(\sigma_i, \mathbf{u}_i, \gamma(\theta))$.
Finally, $\{\mathbf{c}_i\}_{i=1}^S$
are integrated into the final colour
$\mathbf{C}(\mathbf{r})$ using the volumetric rendering equation:
\begin{equation}\label{eq:nerfrendering}
\begin{aligned}
    &\mathbf{C}(\mathbf{r})=\sum_{i=1}^S{w_i\mathbf{c}_i,\,\textrm{ where }w_i=T_i(1-\exp(-\sigma_i\kappa_i})), \\ &T_i=\exp\bigg(-\sum_{j=1}^{i-1}{\sigma_j\kappa_j}\bigg)\textrm{ and }\kappa_i=q_{i+1}-q_i.
    \end{aligned}
\end{equation}

We use hierarchical sampling as in the original NeRF but omit the corresponding notation.

\subsection{Event-based Single  Integral}\label{sec:event_based_integral}
We denote the absolute instantaneous intensity image at time $t$ as $\mathbf{B}(t) \in \mathbb{R}^{W{\times}H}$
and its logarithmic image as
    $\mathbf{L}(t)=\frac{\log \mathbf{B}(t)}{g}$,
where $g$ a gamma correction value fixed to 2.2 in all experiments.
Gamma correction is required to obtain linear colour from $\mathbf{B}(t)$ as it is intended for viewing on the computer screen.
It is encoded with sRGB gamma curve, and  Gamma 2.2 is its recommended smooth approximation~\cite{iec}.
Each event is denoted as a tuple $(t,x,y,p)$, where $t$ is the event timestamp, $(x,y)$ are the 2D coordinates and $p \in \{-1, +1\}$ is the  polarity.
It is assumed to cause a $p\Delta$ change in the logarithmic absolute value $L_{x,y}(t)$ as compared to the value $L_{x,y}(t')$ at the timestamp $t'$ when the previous event happened at $(x,y)$. Hence,
$L_{x,y}(t)-L_{x,y}(t')=p\Delta$,
where $\Delta$ is a fixed and polarity-wise symmetric event threshold.
We model the change in $L_{x,y}(t)$ based on the events as follows:
\begin{equation}
    L_{x,y}(t) - L_{x,y}(t_0) = \int_{t_0}^t \Delta e_{xy}(\tau) d\tau.
\label{eq:Lxy}
\end{equation}
where $t_0$ and $t$ are two different time instants (see Fig.~\ref{fig:method_overview}).
Here, $e_{xy}(\tau)=p\delta_\tau^t$ for all event tuples $(t,x,y,p)$, where $\delta$ is the impulse function with unit integral.
By substituting $e_{xy}(\tau)$ into the integral, the following sum is obtained:
\begin{equation}
    L_{x,y}(t) - L_{x,y}(t_0) = \sum_{(\tau,p): t_0 < \tau \leq t} p\Delta\eqdef E_{x,y}(t_0, t).
    \label{eq:esim}
\end{equation}
The right side of Eq.~\eqref{eq:esim} is a constant computed from the event stream by accumulating per-pixel event polarities.
The left part is the difference between the logarithmic absolute intensity values of the views rendered from $\mathbf{C}(\mathbf{o},\mathbf{d})$ at $t_0$ an $t$.
Thus, Eq.~\eqref{eq:esim} establishes a link between the  observed events $\{\mathbf{E}_i\}_{i=1}^{N}$ and the intensity images.
We propose keeping the right side constant and model $\mathbf{L}$
with the NeRF representation described in Sec.~\ref{sec:NeRF}.
We note that the described model \eqref{eq:esim} is similar to Event-based Double Integral (EDI) model proposed in Pan~\etal~\cite{pan2019bringing}.
However, with NeRF---as opposed to RGB cameras---we can render instantaneous moments, and, hence, there is no motion blur caused by the shutter of such cameras.
Therefore, we reduce Double EDI to Single EDI in the model of Pan~\etal~\cite{pan2019bringing}.  %

\subsection{Self-Supervision with Volumetric Rendering}\label{sec:volume_rendering}

We substitute the left hand side of Eq.~\eqref{eq:esim} by the NeRF-based volumetric rendering $\mathbf{C}$ from Eq.~\eqref{eq:nerfrendering} described in Sec.~\ref{sec:NeRF}.
The differentiable nature of these renderings allows learning
$\mathbf{C}$ of the observed static scene in a self-supervised manner from a single event stream.
We define our reconstruction loss function as the mean square error $\mathbf{MSE(\cdot)}$ of the left side of \eqref{eq:esim} versus its right side:
\begin{equation}
    \mathcal{L}_{\textrm{recon},xy}(t_0, t)=\mathbf{MSE}(\hat{\mathbf{L}}_{xy}(t)-\hat{\mathbf{L}}_{xy}(t_0), E_{xy}(t_0, t)),
    \label{eq:reconloss}
\end{equation}
where $\hat{\mathbf{L}}_{xy}(t)=\frac{1}{g}\log \hat{\mathbf{B}}_{xy}(t)$, with $g$ is a constant gamma correction as discussed earlier.
As per Eq.~\eqref{eq:nerfrendering},
$\hat{\mathbf{B}}_{xy}(t)=\mathbf{C}(\mathbf{r}_{xy}(t))$, where $\mathbf{r}_{xy}(t)$ is the ray corresponding to the pixel at location $(x,y)$.
Note that the derivations concern so far the  grayscale model.

The right-hand argument of Eq.~\eqref{eq:reconloss} is a scalar in case of a grayscale model.
In this case, the $\mathbf{MSE}$ notation should be interpreted in the following way:
The same grayscale value of the right argument
is supervising \textit{all three colour channels}
of the left argument.
This results in learning a grayscale $\hat{\mathbf{L}}(t)$.
The supervision for the case of learning from colour events will be described next.

\subsection{Using Colour Events}
\label{sec:colour}
The difference between colour and grayscale event cameras is the presence of a colour (``Bayer'') filter in front of the sensor~\cite{Rebecq19pami}.
Hence, each pixel perceives one colour channel at the time.
It is known and constant for each pixel and usually it is arranged in $2{\times}2$ blocks.

In traditional camera imaging, all missing colour channel values are recovered
during
\textit{debayering}~\cite{demosaicing}.
For each pixel, its neighbouring values
are used to fill in the missing colour channels.
In event cameras, events fire per pixel asynchronously and,
hence, it is not possible to use traditional debayering.
Another possible solution is to downsample all $2{\times}2$  pixel blocks
into a single pixel,
at the cost of losing $3/4$ of spatial resolution.
We next describe an approach to learn fully coloured model with neither maintaining full frame reconstructions nor losing spatial resolution.

We denote the Bayer filter as $\mathbf{F} \in \mathbb{R}^{W\times H\times 3}$, where $W{\times}{H}$ is the spatial resolution. Here, for each pixel $(x,y)$, $\mathbf{F}_{x,y,z}=1$, only if $z$ is the colour channel of the examined pixel, and $0$ elsewhere.
In the DAVIS 346C event camera which we use in our experiments, $\mathbf{F}$ consists of the following tiled $2{\times}2$ pattern: $[ [[1, 0, 0], [0, 1, 0]], [[0, 1, 0], [0, 0, 1]]]$ (RGGB colour filter).

To model the colour events, we point-wise pre-multiply the colour filter mask $\mathbf{F}$ with both the rendered (left) and the event (right) arguments of the $\mathbf{MSE}$ in Eq.~\eqref{eq:reconloss}:
\begin{equation}
    \mathcal{L}_{\textrm{recon}}(t_0, t)=\mathbf{MSE}(\mathbf{F}\odot \hat{\mathbf{L}}(t)-\mathbf{F}\odot \hat{\mathbf{L}}(t_0), \mathbf{F}\odot E(t_0, t)),
\end{equation}
where ``$\odot$'' denotes pixel-wise multiplication.

In pixels, where only red events happen, the multiplication by $F$ removes all other channels than red.
This does not mean that a particular point in 3D space will only ever be supervised with by the red channel and have no green or blue channel information.
Instead, when the same 3D point is observed from different views, it will be seen by green and blue pixels as well and, hence, have all the needed information to reconstruct the full RGB colour at the full resolution.
Hence we can recover all three RGB channels without demosaicing or loss of spatial resolution of the event windows.

To learn the correct white balance of the scene, we make use of
the assumption that
the background colour $\mathbf{\alpha} \in  \mathbb{R}^{3}$ is known in advance.
The object's silhouette generates events capturing the difference between the background and foreground colours.
This propagates the background colour information into the foreground object, thus recovering its colour as well and
allowing learning both the correct brightness offset and the colour balance between the channels.

\subsection{Final Loss Function}\label{sec:final_loss}

Finally, we obtain the following loss function:
\begin{equation}
\begin{aligned}
    &\mathcal{L}=\frac{1}{N_\mathrm{windows}}\sum_{i=1}^{N_\mathrm{windows}}\mathcal{L}_\mathrm{recon}(t_0, t), \\
    &t =\frac{i}{N_\mathrm{windows}}, t_0\sim U[t-L_\mathrm{max},t),
\end{aligned}
\end{equation}
where $N_\mathrm{windows}=1000$ in our experiments, and the choice of $L_\mathrm{max}$ is described in Sec.~\ref{sec:ray_sampling}.

\subsection{Ray Direction Jitter}
\label{sec:distortion_reg}
An often-used camera trajectory is a circle around the object, so that
the camera keeps roughly the same altitude and distance to the object.
This makes it easy for the model to overfit to the input views.
Following NeRF-OSR~\cite{rudnev2021nerfosr}, we use ray direction jitter for  better generalisation to unseen views, \textit{i.e.,}
instead of shooting the rays exactly through the pixel centre, we randomly offset the ray direction within the pixel.
This policy is fast and easy to implement.

\subsection{Event-based Ray Sampling Strategy}
\label{sec:ray_sampling}
Our method can be used with individual events or event  windows; we use event windows for efficiency and for simplicity.
Thus, we accumulate events from the start to the end of the window.
Now we describe how we select the starts and the ends of the windows.
First, we split the whole event stream into intervals  $t_i=i/N_\textrm{windows}$, where
$i \in \{1, \hdots, N_\textrm{windows}\}$;
$t_i$ are the window ends.
We empirically noticed that using constant short windows results in poor propagation of high-level lighting;
using constant long windows often results in poor local details.
Hence, we sample the window length randomly as $t-t_0{\sim}U(0, L_\mathrm{max}]$.
Equivalently, if the window end $t$ is fixed, the window start is sampled as $t_0{\sim}U[t-L_\mathrm{max},t)$.
This allows the model to learn both global and local colour information.
For real data experiments, we use $L_\mathrm{max}=0.1$, where $t=1.0$ is the end of the stream.
For synthetic data experiments, we use $L_\mathrm{max}=0.05$.

In every event window, $N_e$ pixels are sampled from the positions where $E(t_0, t)\neq 0$, \ie, at least one event happens and is not cancelled by others.
However, if sampling is limited only by non-zero rays (\emph{positive sampling}),
information about pixels with no events will be ignored and would degrade results.
Hence, we also apply \emph{negative sampling}:
We sample $\beta N_{e}$ rays from every other pixel in the frame where $\beta=0.1$.
This makes our method more robust to noise events
and leads to better reconstruction of the uniformly coloured surfaces
as shown in the ablation studies (Sec.~\ref{sec:ablation}).
It also helps in reconstructing thin structures.

In total, our sampling strategy uses only $O((1+\beta)N_{e})=O(N_{e})$ rays per epoch.
This is much more efficient than the ray sampling strategy used in traditional NeRF for processing RGB images, where in each epoch, every pixel is rendered from every camera viewpoint, thus requiring  $O(W{\times}{H}{\times}{N}_\mathrm{views})$ rays, where $W{\times}{H}$ is the resolution and ${N}_\mathrm{views}$ the number of camera viewpoints.
This observation proposes to discard the traditional notion of training view count~\cite{Tewari2022NeuRendSTAR} in our work and to instead use new data-efficiency metrics based on the number of events.
It also suggests that redundant views do not help model learning as much as when they were diversely sampled.

\subsection{Implementation  Details}\label{ssec:implementation_details}
Our code is based on NeRF++~\cite{kaizhang2020} without
the background network as our test scenes can completely fit inside the unit sphere rendered using the foreground network.
We replace $\mathrm{ReLU}$ with $\tanh$ activation function as
it results in more consistent training results.
We train the models for $5\cdot 10^5$ iterations on two NVIDIA Quadro RTX 8000 GPUs with the optimiser and hyper-parameters as in NeRF++.
This takes about six hours to complete.
Rendering a full $346{\times}260$ image (\textit{i.e.,} the resolution of DAVIS~346C) then takes about $1.7$ seconds using the same hardware setup.
Our real-time demo is based on torch-ngp~\cite{torch-ngp} (see the supplement).
We train a model for $5\cdot 10^3$ iterations on a single NVIDIA RTX 2070, which takes around one minute to converge.
After training, rendering in Full HD resolution is real-time at $30$ FPS using the same hardware setup.

\section{Experiments}\label{sec:experiments}

\begin{figure}
    \centering
    \footnotesize
    \includegraphics[width=0.98\linewidth]{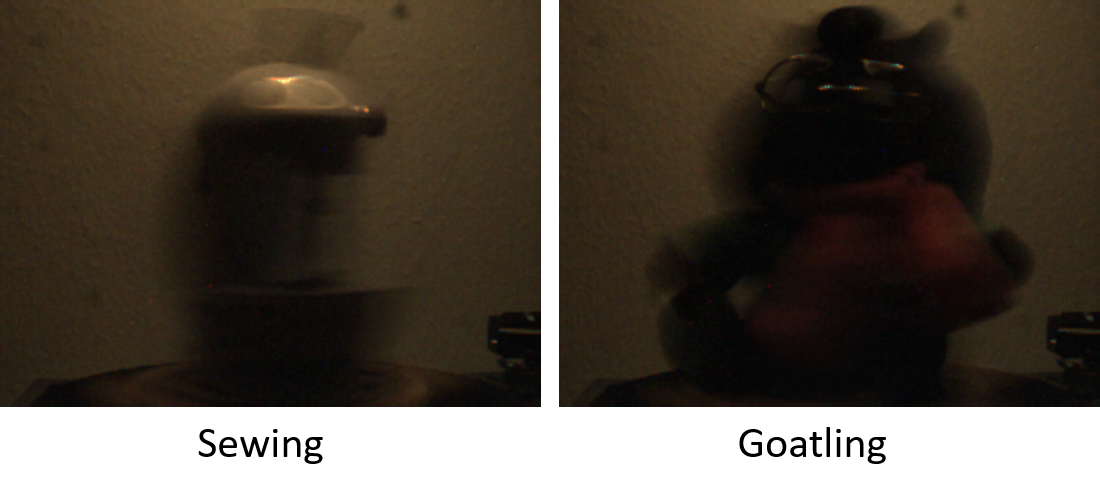}
    \vspace{-10pt}
    \caption{Showing the fact that our real sequences were recorded under low lighting conditions. Here, we show the RGB stream of the event camera that was used for recording the real sequences. These sequences were recorded using just a 5W USB light source. We tone map the real sequences in the remaining figures for clarity.}
    \label{fig:lowlight}
\end{figure}

\begin{figure*}
    \centering
    \includegraphics[width=1.0\textwidth]{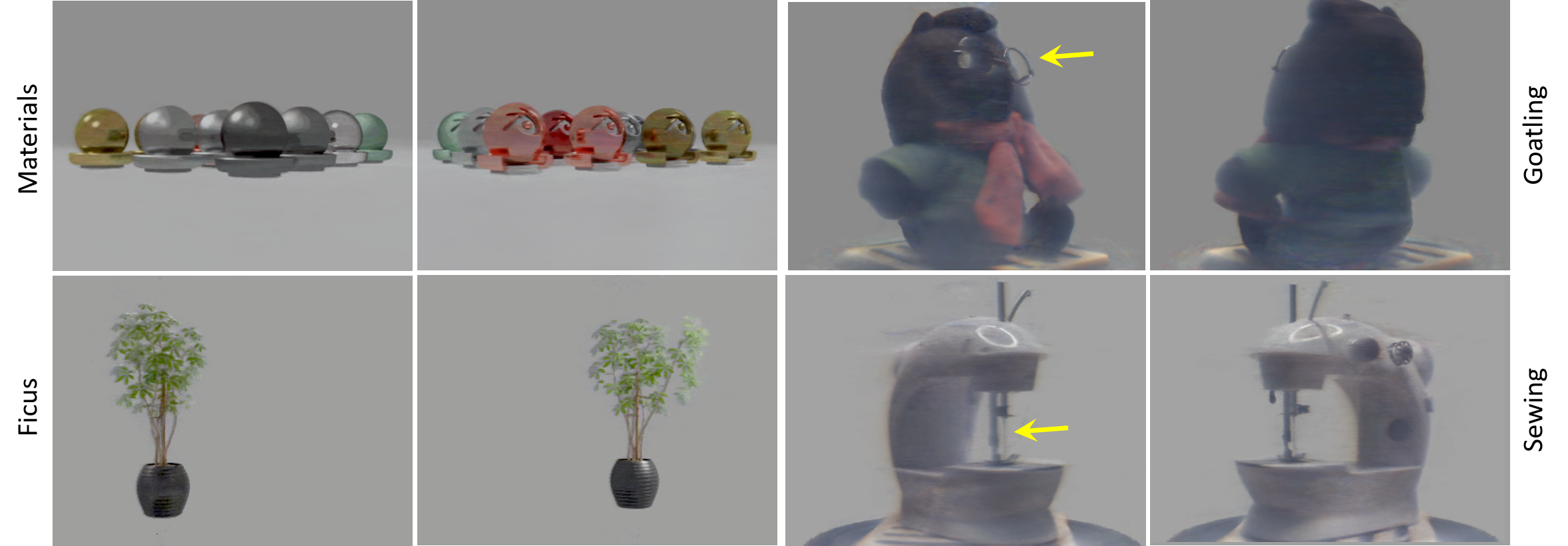}
    \caption{Reconstructions by our EventNeRF on synthetic (``Materials'', ``Ficus'') and real (``Goatling'', ``Sewing'') sequences. Our approach reconstructs specularities (``Materials''), thin structure (``Ficus'') and even the eyeglass frame and the needle from the ``Goatling'' and ``Sewing'' sequences respectively (see yellow arrow).
    }
    \label{fig:SynReal}
\end{figure*}

\noindent\textbf{Examined Sequences.} We examine seven synthetic and ten real sequences.
As synthetic ones, we use the 3D models from  Mildenhall~\etal~\cite{mildenhall2020nerf}.
For each scene, we render a one-second-long $360^\circ$ rotation of camera around the object at 1000~fps as RGB images, resulting in 1000 views.
From these images, we simulate the events stream using the model from~\cite{rudnev2021eventhands}.
The corresponding camera intrinsics and extrinsics are used directly in our method.
For the real-data experiments, we record ten objects with the DAVIS~346C colour event camera on a uniform white background.
We notice that it is hard to make a stable and calibrated setup in real life where the event camera rotates around an object.
In the absence of the background events, rotating the object while keeping the camera still results in the same event stream as when the camera rotates around the static object.
Hence, we keep the camera static and place the objects on a direct drive vinyl turntable, resulting in stable and consistent object rotation at 45~RPM.
In this setting, it is essential to provide constant lighting
regardless of the current object's rotation angle, and
we mount a single USB ring light above the object.
A photo of the setup is included in the supplement (Fig.~1).
Please note that all the real sequences are captured under low lighting condition (see Fig.~\ref{fig:lowlight}), as the light source we use have a power of just 5W.

\textbf{Metrics.} To account for the varying event generation thresholds, colour balance, and exposure, we do the following procedure for all our and baseline results.
In our model, the event threshold only affects the overall image contrast in logarithmic space.
Exposure and colour balance are modelled as an offset in logarithmic space.
Hence for every sequence of predicted images ${\mathbf{I}_{k}}_{k=1}^{N_\mathrm{images}}$ and the corresponding ground-truth images ${\mathbf{G}_{k}}_{k=1}^{N_\mathrm{images}}$, we fit a single linear colour transform $f(\mathbf{I})$, which we apply to the predictions in the logarithmic space:
\begin{equation}
\small
\begin{aligned}
    &f(\mathbf{I}) = \exp(\mathbf{a}\odot \log\mathbf{I}+\mathbf{b}), \\
    &(\mathbf{a}, \mathbf{b}) = \arg\min_{\mathbf{a}, \mathbf{b} \in \mathbb{R}^3}
    \sum_{k=1}^{N_\mathrm{images}}
    \|\mathbf{a}\odot\log\mathbf{I}_{k}+\mathbf{b}-\log\mathbf{G}_{k}\|^2.
\end{aligned}
\end{equation}

Then, for the transformed images ${f(\mathbf{I}_{k})}_{k=1}^{N_\mathrm{images}}$, we
report PSNR, SSIM, and LPIPS~\cite{zhang2018perceptual} based on AlexNet.

\subsection{Synthetic Sequences}

As described earlier, we examine synthetic sequences from Mildenhall~\etal~\cite{mildenhall2020nerf}.
They cover
different
effects such as thin structures (drums, caterpillar, ficus, microphone), view-dependent effects (drums, caterpillar, materials), large uniformly coloured surfaces (chair).
Fig.~\ref{fig:SynReal}-(left) shows visual results for two sequences.
EventNeRF learns view-dependent effects (Materials) and thin structures (Ficus).
Fig.~\ref{fig:E2VID} shows our reconstruction on two more synthetic sequences, Here, we capture textured regions well (Lego).
Corresponding numerical results are reported in Tab.~\ref{tbl:e2vid}.

\subsection{Real Sequences}

We examine ten scenes recorded on a turntable: Goatling, dragon, chicken, sewing machine, game controller, cube, bottle, fake plant, multimeter and microphone.
They show the performance of our method on scenes with thin structures (goatling, sewing machine, fake plant, microphone), fine-print coloured text (game controller, multimeter, bottle), view-dependent effects (goatling, sewing machine, cube, bottle), dark details (goatling, game controller).
As there is no ground-truth RGB data, we can only evaluate results visually, as shown in Figs.~\ref{fig:SynReal} and~\ref{fig:E2VID-real}.
Note how can we can reconstruct a one-pixel-wide sewing needle in the ``Sewing'' sequence, and
the eyeglasses frame on the ``Goatling'' scene (see yellow arrows in Fig.~\ref{fig:SynReal}).
Our results are also halo-free despite being reconstructed from events only.
In Fig.~\ref{fig:depthmaps}, we show the extracted depth. Here, we use a colour scheme that represents per-pixel distances in meters from a virtual camera to the object in a novel view.

\subsection{Comparisons against Related Methods}

\begin{figure}
    \centering
    \footnotesize
    \includegraphics[width=\linewidth]{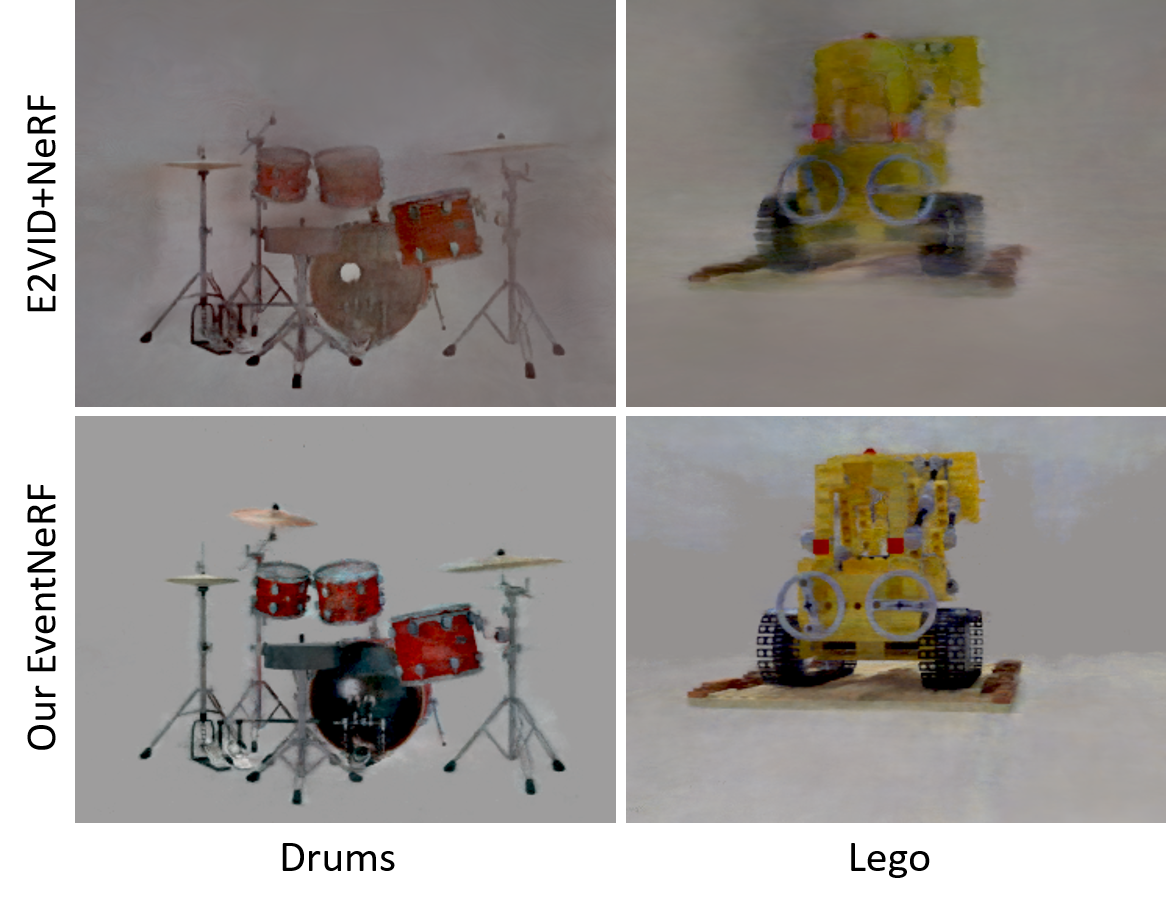}
    \vspace{-15pt}
    \caption{
    EventNeRF better resembles the ground truth than the baseline E2VID+NeRF.
    }
    \label{fig:E2VID}
\end{figure}

\begin{table}
\scriptsize
    \centering
\resizebox{\columnwidth}{!}{%
\begin{tabular}{c|c|c|c|c|c|c}
 & \multicolumn{3}{c|}{E2VID \cite{Rebecq19pami} + NeRF \cite{mildenhall2020nerf}} & \multicolumn{3}{c}{Our EventNeRF}\\
Scene & PSNR$\uparrow$ & SSIM$\uparrow$ & LPIPS$\downarrow$ & PSNR$\uparrow$ & SSIM$\uparrow$ & LPIPS$\downarrow$ \\
\hline
Drums & $19.71$ & $0.85$ & $0.22$ & $\mathbf{27.43}$ & $\mathbf{0.91}$ & $\mathbf{0.07}$\\
Lego & $20.17$ & $0.82$ & $0.24$ & $\mathbf{25.84}$ & $\mathbf{0.89}$ & $\mathbf{0.13}$\\
Chair & $24.12$ & $0.92$ & $0.12$ & $\mathbf{30.62}$ & $\mathbf{0.94}$ & $\mathbf{0.05}$\\
Ficus & $24.97$ & $0.92$ & $0.10$ & $\mathbf{31.94}$ & $\mathbf{0.94}$ & $\mathbf{0.05}$\\
Mic & $23.08$ & $0.94$ & $0.09$ & $\mathbf{31.78}$ & $\mathbf{0.96}$ & $\mathbf{0.03}$\\
Hotdog & $24.38$ & $0.93$ & $0.12$ & $\mathbf{30.26}$ & $\mathbf{0.94}$ & $\mathbf{0.04}$\\
Materials & $22.01$ & $0.92$ & $0.13$ & $\mathbf{24.10}$ & $\mathbf{0.94}$ & $\mathbf{0.07}$\\
\hline
Average & $22.64$ & $0.90$ & $0.15$ & $\mathbf{28.85}$ & $\mathbf{0.93}$ & $\mathbf{0.06}$\\
\end{tabular}
}
\caption{Comparing our method against E2VID+NeRF. Our method  consistently produces better results.
}
    \label{tbl:e2vid}
\end{table}

\begin{figure}
    \centering
    \footnotesize
    \includegraphics[width=\linewidth]{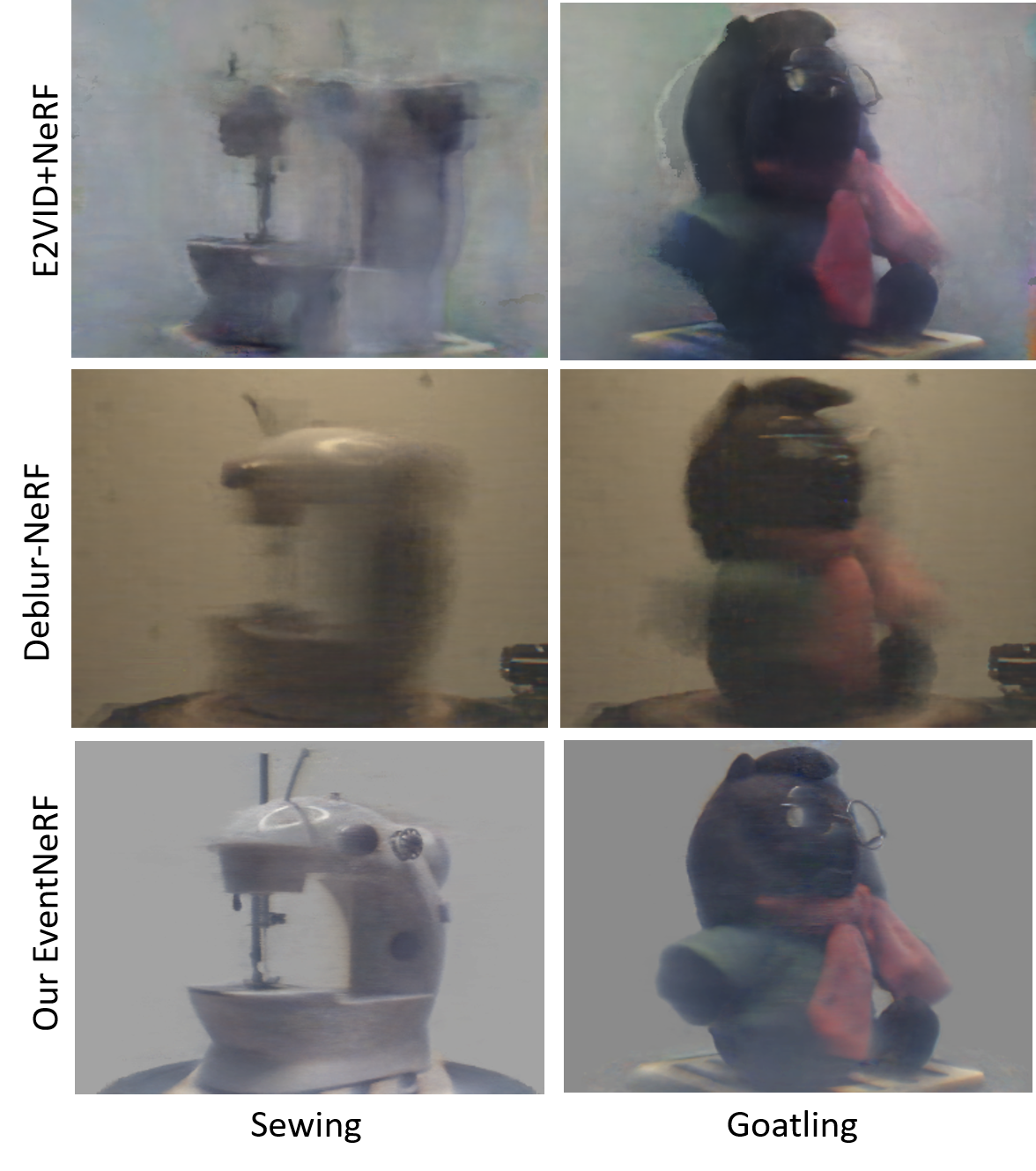}
    \vspace{-17pt}
    \caption{
    On real data our EventNeRF clearly outperforms all the related methods of E2VID+NeRF and Deblur-NeRF.
    }
    \label{fig:E2VID-real}
\end{figure}

We first recover RGB frames from events using E2VID~\cite{Rebecq19pami} and then learn original frame-based NeRF with them as the training data;
see Tab.~\ref{tbl:e2vid}.
EventNeRF clearly outperforms this approach (which we call E2VID+NeRF) in all metrics and on all sequences.
Fig.~\ref{fig:E2VID} shows visual samples of this experiment on the Drums and Lego sequences.
Fig.~\ref{fig:E2VID-real} show that E2VID+NeRF also generates noticeable artefacts on the real sequences.
Note that replacing E2VID with ssl-E2VID~\cite{Paredes21} generates even worse results, as the performance of ssl-E2VID is usually bounded by E2VID (see the supplement).
Note that such approaches do not account for the sparsity and asynchronousity of the event stream
and need much more memory and disk space to store all the reconstructed views.
Moreover, there is a limit to how short the E2VID window can be made, and hence to the number of reconstructed views.
In contrast, EventNeRF respects the asynchronous nature of event streams and reconstructs the neural representation directly from them; it can use an arbitrary number of windows, which allows reconstructing the scene even from $3\%$ of the data, as we show in Sec.~\ref{sec:ablation}.
That makes it significantly more scalable than first reconstructing the frames and using traditional NeRF \cite{mildenhall2020nerf}.

We also compare against Deblur-NeRF~\cite{ma2021deblur}, \textit{i.e.,} a RGB-based NeRF extraction method designed specifically to handle blurry RGB videos.
We perform this comparison by applying Deblur-NeRF on the the RGB stream of the event camera.
Results in Fig.~\ref{fig:E2VID-real}, however, show that our approach significantly outperform Deblur-NeRF.
This is expected, as Deblur-NeRF can only handle view-inconsistent blur.
On the other hand, our approach produces almost blur-free results.
In addition, it is significantly more memory and computationally efficient.
For Deblur-NeRF, $100$ training views were used, which took around 22 seconds to record due to the low-light condition (see Fig.~\ref{fig:lowlight}).
Our EventNeRF approach, however, need only one $1.33$s revolution of the object.
In addition, it converges well within 6 hours using two NVIDIA Quadro RTX 8000 GPUs.
This compares favourably to Deblur-NeRF, which needs $16$ hours with the same GPU resources.
Furthermore, the training time for our method significantly drops with the torch-ngp~\cite{torch-ngp} implementation to just 1 minute using a single NVIDIA GeForce RTX 2070 GPU.
This, however, comes with some compromise in the rendering quality (see the supplemental video).
We also evaluated Deblur-NeRF on two synthetic sequences where we simulated the blur to match real recorded sequences.
Our technique performs significantly better in PSNR, SSIM and LPIPS:
$(27.43,0.91,0.07)$ for Drums and $(25.84,0.89,0.13)$ for Lego; and Deblur-NeRF scores much worse at $(21.61,0.76,0.36)$ and $(21.06,0.76,0.35)$.

\begin{figure}
    \centering
    \footnotesize
    \includegraphics[width=\linewidth]{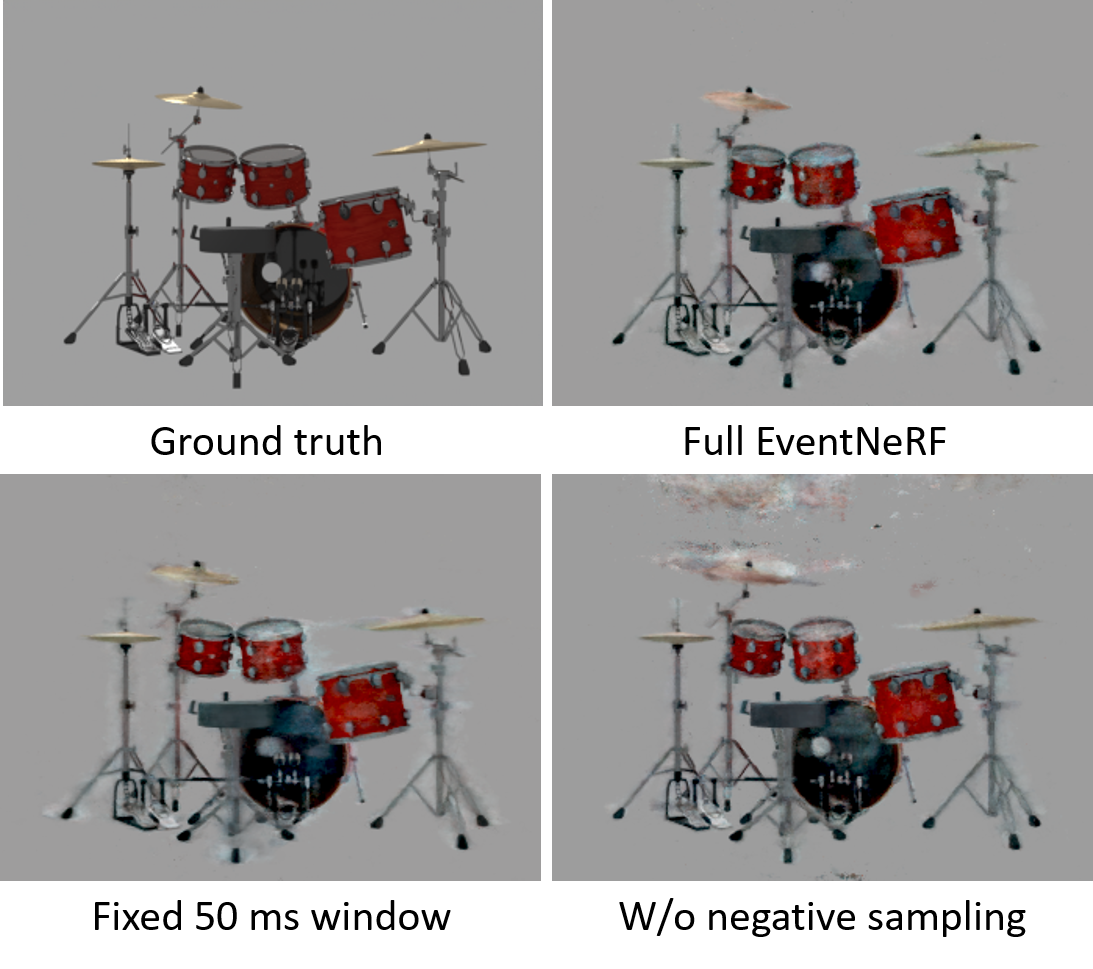}
    \vspace{-15pt}
    \caption{Importance of our various design choices (see Sec.~\ref{sec:ablation}). The full model produces the best results.
    }
    \label{fig:ablation}
\end{figure}

\begin{table}
    \centering
    \begin{tabular}{c|c|c|c}
    Method & PSNR $\uparrow$ & SSIM $\uparrow$ & LPIPS $\downarrow$ \\
    \hline
    Fixed 50~ms win. & $27.32$ & $0.90$ & $0.09$\\
    W/o neg. smpl. & $26.48$ & $0.87$ & $0.16$\\
    \hline
    Full EventNeRF & $\mathbf{27.43}$ & $\mathbf{0.91}$ & $\mathbf{0.07}$\\
    \end{tabular}
    \caption{Ablation studies computed on the \textit{Drums}.
    }
    \label{tab:ablation}
\end{table}

Our results on real data show that EventNeRF handles poorly lit environments (all our sequences were shot in dim lighting; see Fig.~\ref{fig:lowlight}).
Note that we cannot compare against RAW-NeRF~\cite{mildenhall2021rawnerf} and HDR-NeRF~\cite{huang2021hdrnerf} as they require fundamentally different setups.
RAW-NeRF~\cite{mildenhall2021rawnerf} requires the RAW data as input and HDR-NeRF~\cite{huang2021hdrnerf} requires footage of the same object shot with different exposures.
Hence, they cannot be applied to
the RGB frames of an event camera.

\subsection{Ablation Study and Data Efficiency}
\label{sec:ablation}
We ablate the use of negative sampling and analyse window size randomisation as an alternative to a fixed long window.
The results are shown both in Tab.~\ref{tab:ablation} and in Fig.~\ref{fig:ablation} on ``Drums''.
The most significant degradation comes from disabling negative sampling, as this leads to the highly visible artefacts in the background and from the object.
Furthermore, using fixed 50~ms window results in the loss of fine detail and the emergence of halo in the reconstruction.

\begin{figure}
    \centering
    \includegraphics[width=\linewidth]{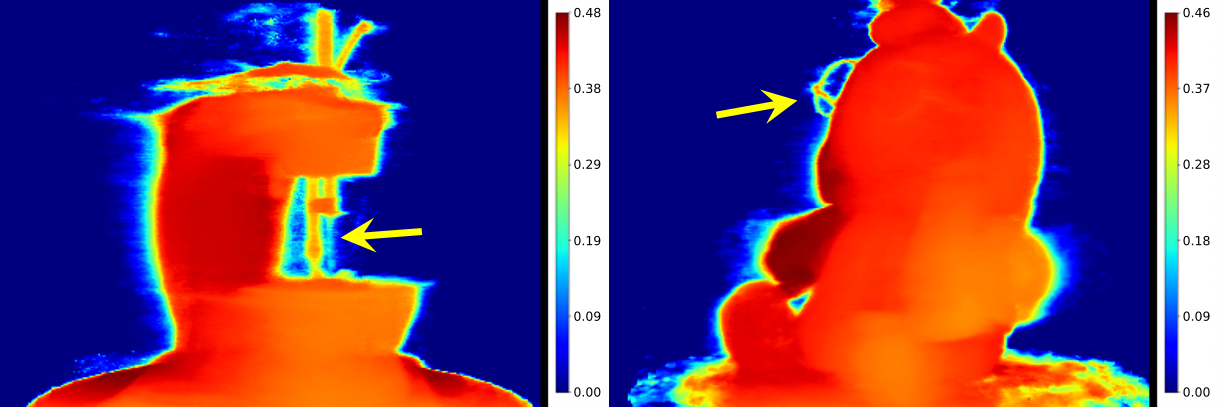}
    \caption{Examples of depth maps (in meters) at arbitrary novel views obtained in our tests with EventNeRF with real data (left: ``Sewing'', right: ``Goatling''). Note the fine details such as the sewing needle and the eyeglasses frame highlighted by the arrows. 
    Rays that do not hit the object have zero depth. 
   }
    \label{fig:depthmaps}
\end{figure}

\begin{figure}
    \centering
    \includegraphics[width=\linewidth]{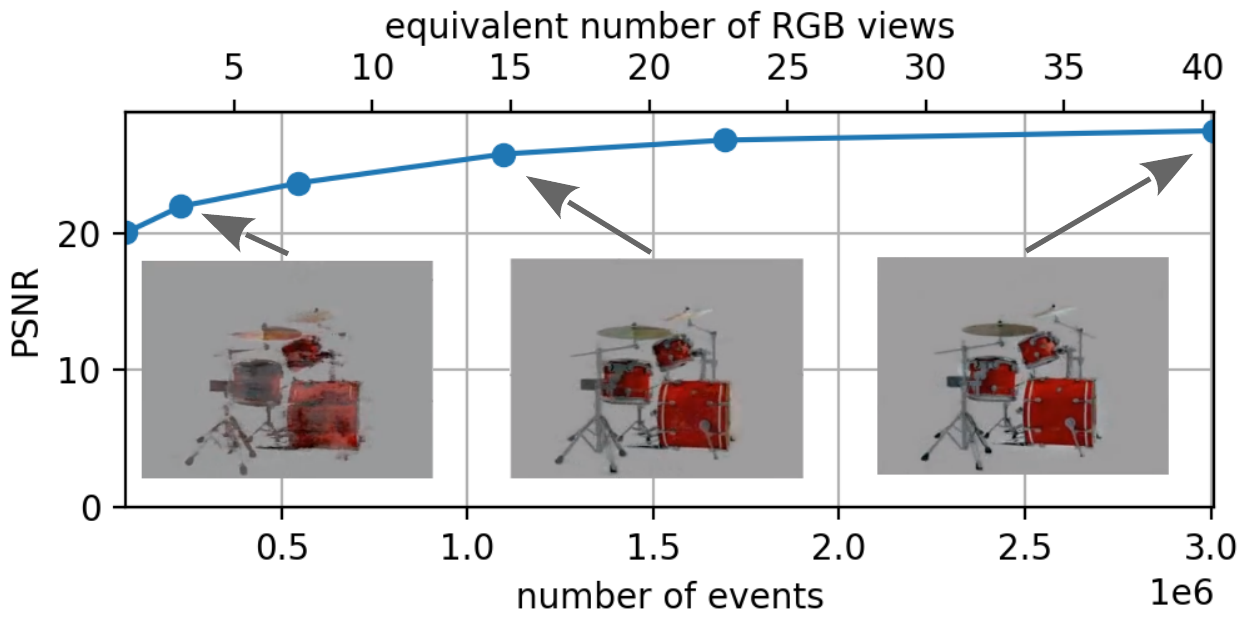}
    \vspace{-17pt}
    \caption{PSNR performance of our EventNeRF on ``Drums'' with varying event numbers (multiplied by $10^6$ on the $x$-axis). The images underneath the function graph show exemplary novel views obtained with the $7\%$, $36\%$ and $100\%$ of the events, respectively.
    }
    \label{fig:efficiency}
\end{figure}

By varying the event generation threshold $\Delta$ in the simulated event streams, we can vary the total number of events.
With higher $\Delta$, the simulated camera becomes less sensitive to brightness changes in a fashion similar to quantisation.
At a high enough $\Delta$, some details in the scene never trigger a single event.
Hence, we can compare how our method performs with different event numbers.
In the Fig.~\ref{fig:efficiency}, we plot PSNR of the reconstructed ``Drums'' scene as a function of the number of events and the number of equivalent RGB frames (that would occupy the same space).
The generated novel views with $36\%$ and $100\%$ of the available events are barely distinguishable and with $7\%$ of the data, the scene is still well recognisable.
Note how we can still reconstruct the scene with the amount of data that is less than \textbf{one} equivalent RGB frame at 20 PSNR.
For reference, NeRF's performance degrades significantly with less than 100 views~\cite{dietnerf}.

\section{Conclusion}
We introduced the first method to reconstruct a 3D model of a static scene from event streams that enables dense photorealistic RGB view synthesis.
Thanks to the proposed combination of event supervision, volumetric rendering and several event-specific regularising techniques, EventNeRF outperforms
the baselines in the rendered image quality, fast motion handling, low-illumination handling, and data storage requirements.
Thus, this paper extends the spectrum of practical event-based techniques with a 3D representation learning approach.
Future work can investigate joint estimation of the camera parameters and the NeRF volume.

\noindent\textbf{Acknowledgements.} 
This work was supported by the ERC  Consolidator Grant \textit{4DReply}  (770784).

{\small
\bibliographystyle{ieee_fullname}
\bibliography{egbib}
}

\appendix
\clearpage
\begin{center}
\textbf{\Large Appendices}
\end{center}

This appendix provides more detail on the experiments and technical aspects for the proposed technique.
In Sec.~\ref{sec:realdatacapture} we discuss our experimental setup for capturing real data and describe our camera calibration and other details.
We then show more results on synthetic (Sec.~\ref{sec:synthetic}) and real data (Sec.~\ref{sec:real}).
Here, we show visual results for ssl-E2VID~\cite{Paredes21}, which as mentioned in the main text are clearly worse than E2VID~\cite{Rebecq19pami}.
We then show an application of running our approach in real-time through an instant-ngp implementation~\cite{torch-ngp} (Sec.~\ref{sec:realtime}).
We demonstrate the ability to extract meshes from our trained models in Sec.~\ref{sec:mesh}.
Then we provide details on the window sampling in Sec.~\ref{sec:win_sampling}. 
Finally, we include an additional ablation study for our method, \textit{i.e.,} on real data (Sec.~\ref{sec:ablation_real}) and study the effect of inaccurate camera poses (Sec.~\ref{sec:abl_angle}) and the robustness to noise events in the training data  (Sec.~\ref{sec:abl_noise}). 

\begin{figure}[h]
    \centering
    \includegraphics[height=7.25cm]{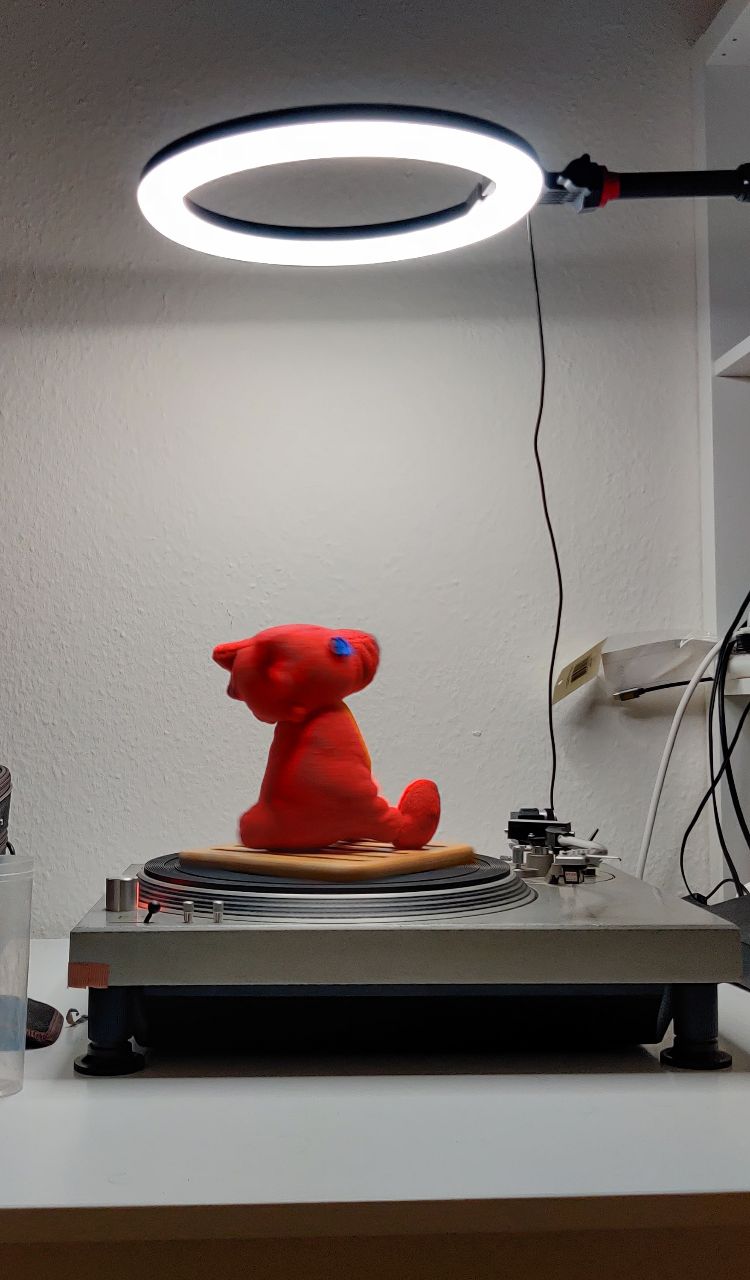}
    \includegraphics[height=7.25cm]{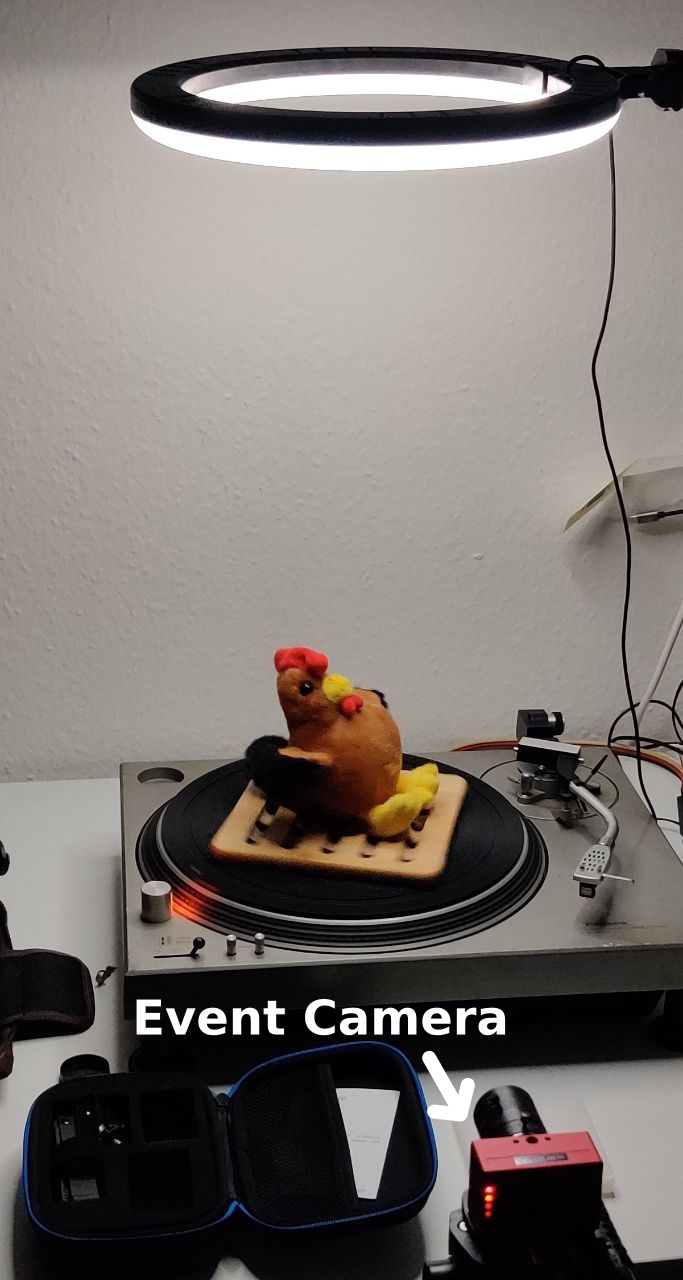}
    \caption{Our real data recording setup. The object is placed on a 45 RPM direct-drive vinyl turntable and lit by a 5W USB ring light mounted directly above it. The scene is recorded with a DAVIS 346C colour event camera (right bottom).
    }
    \label{fig:setup}
\end{figure}

\section{Real Data Capture}
\label{sec:realdatacapture}

We use the DAVIS 346C colour event camera to record our real sequence. Fig.~\ref{fig:setup} shows photos of the setup we used to record the real data. We used the default camera settings in the DV software provided with the camera.

\subsection{Camera Pose Calibration}
\label{sec:cpc}

We estimate the extrinsic parameters of the event camera as follows. 
We noticed that due to the constant rotation speed of the turntable, camera extrinsics can be computed analytically as a point uniformly moving on a circle looking at its centre.
In practice, this requires both precise mechanical and computational calibration. 

\begin{figure}[h]
    \centering
    \includegraphics[height=4.5cm]{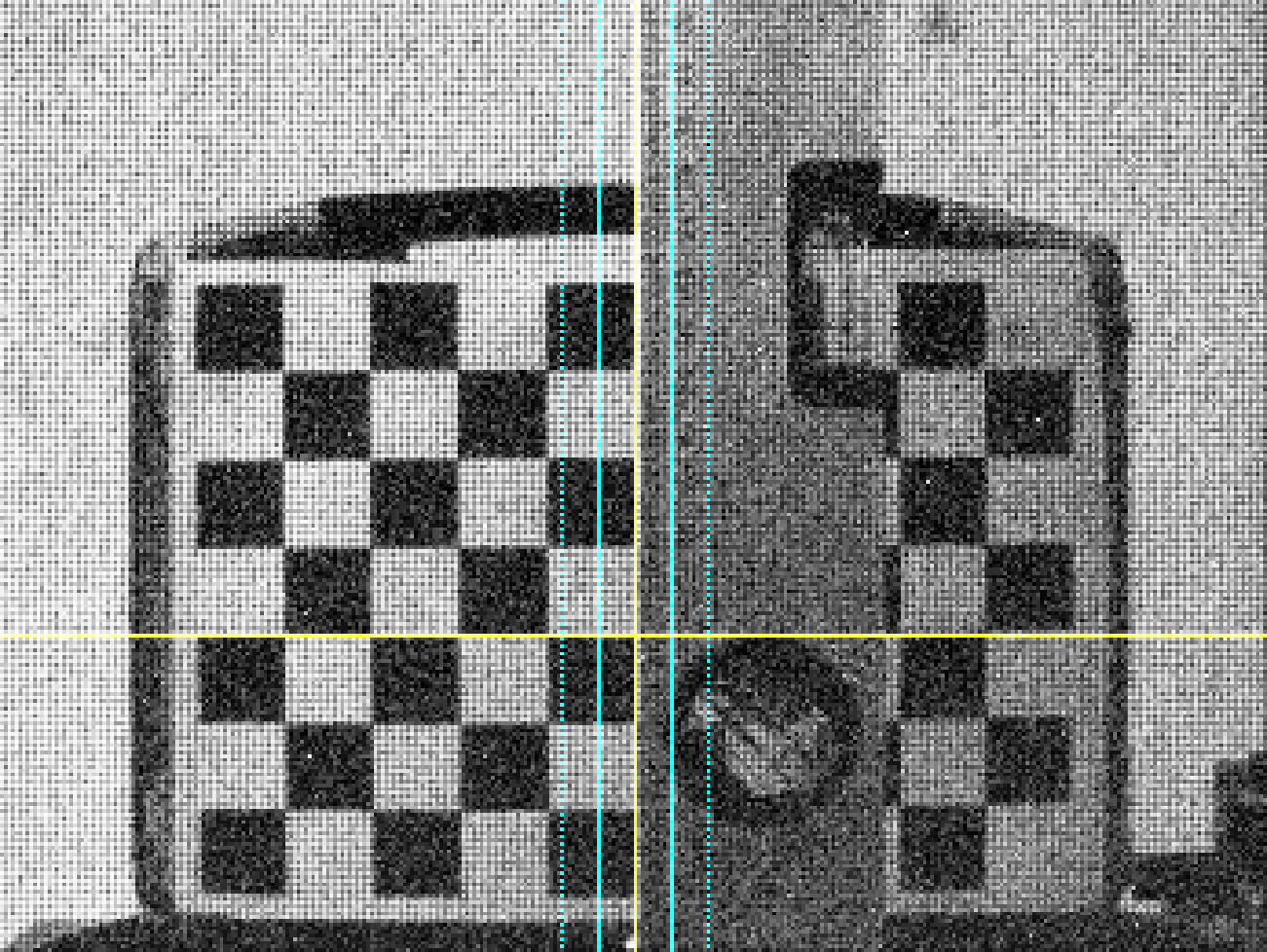}
    \caption{Results of the mechanical adjustment of the camera tripod as seen through the camera itself. The chequerboard and the ruler are placed exactly at the turntable rotation axis. Here, yellow is the vertical going through the optical centre of the camera. As seen in the picture, both axes match up to a pixel. Please note that the blue lines mark +/-10 and +/-20 pixels from the centre.
    }
    \label{fig:calib_match}
\end{figure}

First, we adjusted the camera tripod as precise as possible so that the vertical through the optical centre of the camera matches the turntable rotation axis (Fig.~\ref{fig:calib_match})

\begin{figure}[h]
    \centering
    \includegraphics[height=7.25cm]{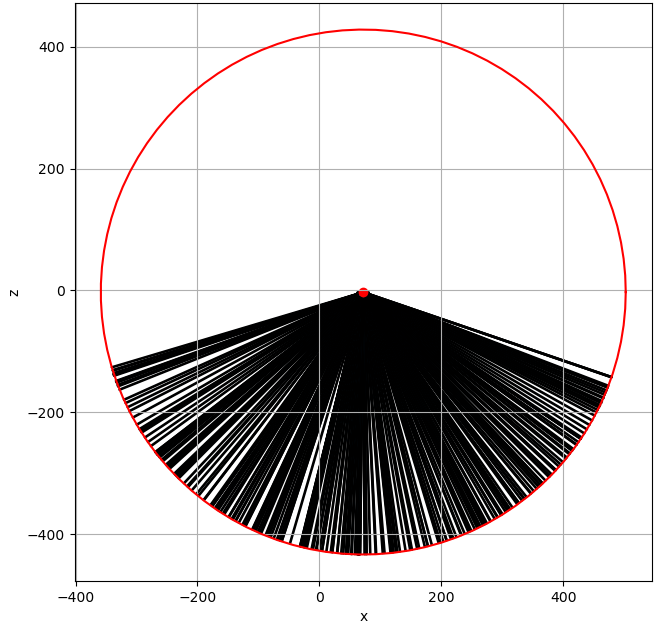}
    \includegraphics[width=\columnwidth]{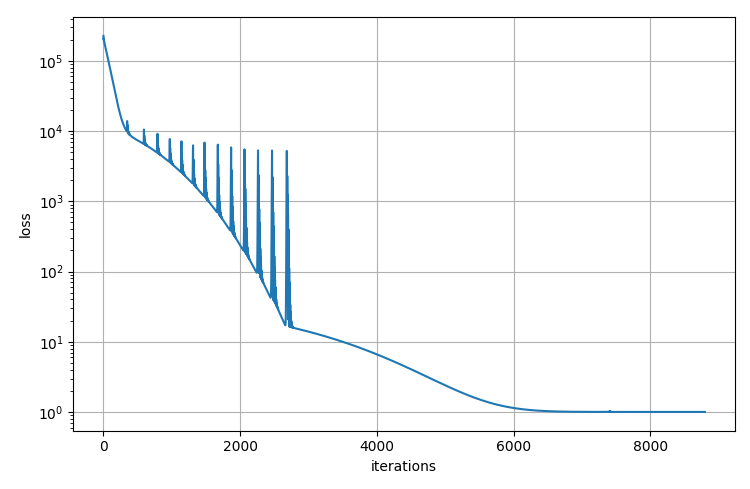}
    \caption{Camera extrinsic optimisation results (top) and convergence process (bottom) using the chequerboard protocol. In red is the optimised circle and the circle centre. In black are the arrows starting from the extracted camera poses meeting in the circle centre. In blue is the loss function convergence; the peaks in the optimisation plot are caused by learning rate scheduling. 
    }
    \label{fig:calib_optim}
\end{figure}

Second, we estimated the residual offset using the following protocol.
We placed the chequerboard pattern with its centre as close on the turntable's axis of rotation as possible.
Then we slowly rotated the plate and recorded corresponding RGB frames using the event camera.
Using this data, we found the positions and rotations of the camera in space wrt. the chequerboard.
These positions lie on the circle, which corresponds to the correct camera poses, and they are tilted to the rotational axis with an unknown angle offset $\alpha$.
We solve for $\alpha$, circle radius and circle centre coordinates via optimisation.
The optimisation objective is such that all the rays coming through the optical centre of the cameras must meet in the same point in space that is the centre of the circle at a distance that equals to the radius of the circle.
We use Adam~\cite{2015KingmaBa} optimiser with its learning rate reduced on plateaus. 
We show the converged results and convergence process on Fig.~\ref{fig:calib_optim}.
In our recordings, we found that
$\alpha=2.85^\circ$ for the Goatling and Sewing recordings and $\alpha=0.2388^\circ$ for the rest of the sequences.

\subsection{Density Clipping}
For the real scenes, we know that the object always lies inside the cylinder defined by the turntable plate.
Hence, to filter the noise and artefacts in the unobserved areas, we force the density to zero everywhere outside of this cylinder:
\begin{equation}
    \sigma(x,y,z)=0\textrm{, if }x^2+y^2>r_\mathrm{max}^2\textrm{ or }z>z_\mathrm{max}\textrm{ or }z < z_\mathrm{min}.
\end{equation}
The cylinder parameters $z_\mathrm{min}$, $z_\mathrm{max}$ and $r_\mathrm{max}$ are tuned manually to fit the recorded experimental setup.
In our case,
 $z_\mathrm{min}=-0.35$, $z_\mathrm{max}=0.15$ and $r_\mathrm{max}=0.25$.

\section{Synthetic Data Results}
\label{sec:synthetic}
We provide more synthetic data results and comparisons in Fig.~\ref{fig:moresynthetic} and in the supplementary video.
E2VID~\cite{Rebecq19pami}+NeRF~\cite{mildenhall2020nerf} struggles with background reproduction and separation, resulting in less clear images and background artefacts.
In addition, the colour and detail reproduction are also a concern for E2VID+NeRF.
Our method does not suffer from such problems. It produces photorealistic results, capturing specularities (Lego, Materials, Microphone), thin structures (Drums, Ficus, Lego, Microphone), and textured regions (Hotdog, Chair).

Note that Tab.~\ref{tbl:e2vid} shows that our approach clearly outperforms E2VID~\cite{Rebecq19pami}+NeRF~\cite{mildenhall2020nerf}.

\begin{figure*}[ht]
    \centering
    \begin{tabular}{@{}c@{\hspace{0.05cm}}c@{\hspace{0.05cm}}c@{\hspace{0.05cm}}c@{\hspace{0.05cm}}c@{\hspace{0.05cm}}c@{}}
    \includegraphics[height=2.15cm, clip=true]{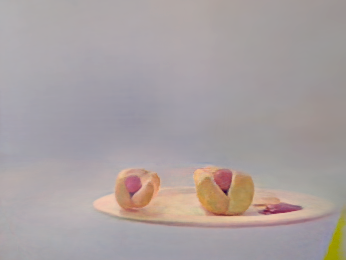}&
    \includegraphics[height=2.15cm, clip=true]{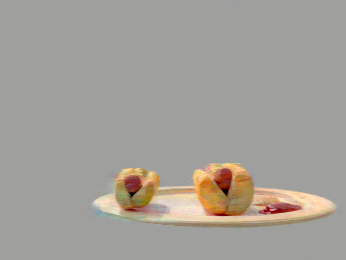}&
    \includegraphics[height=2.15cm, clip=true]{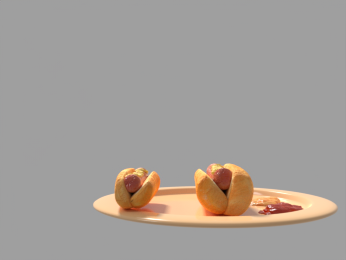}&
    \includegraphics[height=2.15cm, clip=true]{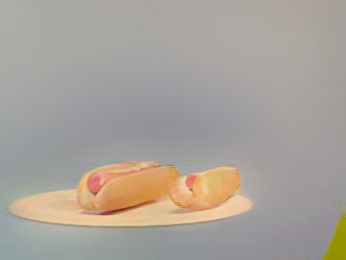}&
    \includegraphics[height=2.15cm, clip=true]{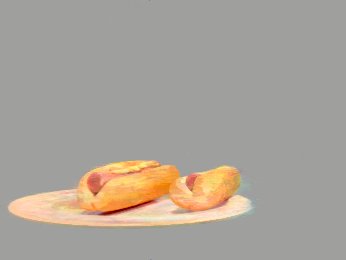}&
    \includegraphics[height=2.15cm, clip=true]{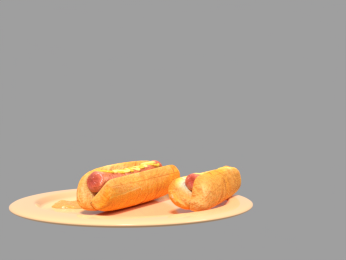}\\
    \includegraphics[height=2.15cm, clip=true]{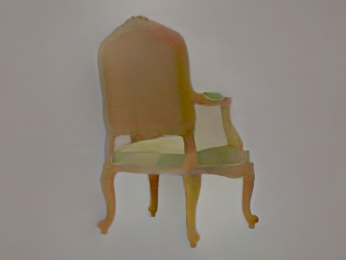}&
    \includegraphics[height=2.15cm, clip=true]{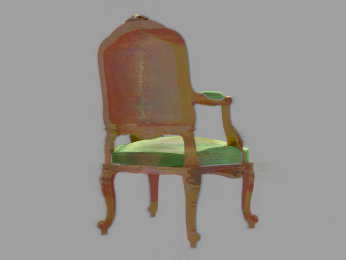}&
    \includegraphics[height=2.15cm, clip=true]{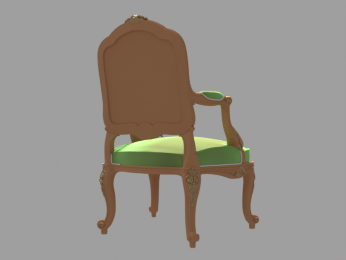}&
    \includegraphics[height=2.15cm, clip=true]{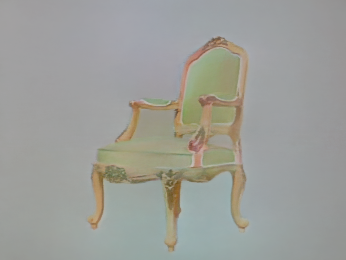}&
    \includegraphics[height=2.15cm, clip=true]{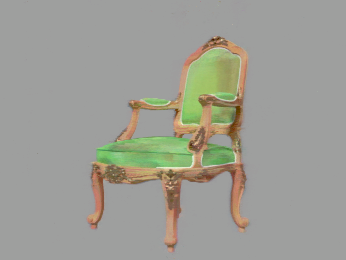}&
    \includegraphics[height=2.15cm, clip=true]{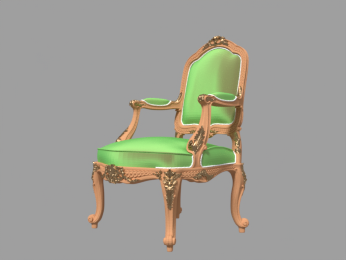}\\
    \includegraphics[height=2.15cm, clip=true]{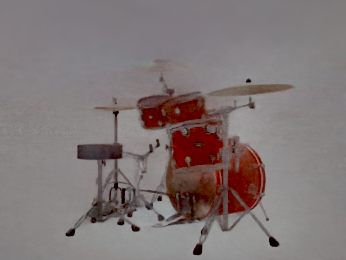}&
    \includegraphics[height=2.15cm, clip=true]{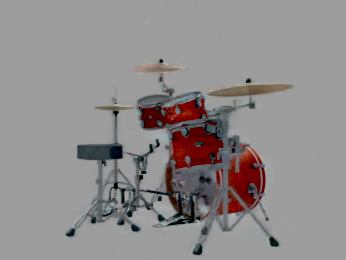}&
    \includegraphics[height=2.15cm, clip=true]{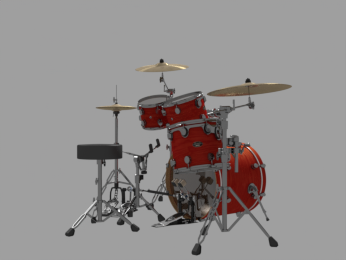}&
    \includegraphics[height=2.15cm, clip=true]{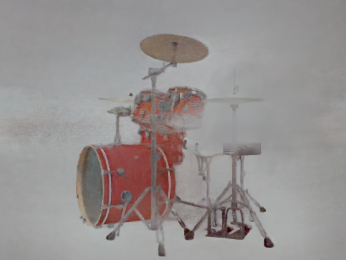}&
    \includegraphics[height=2.15cm, clip=true]{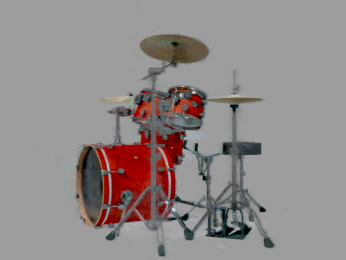}&
    \includegraphics[height=2.15cm, clip=true]{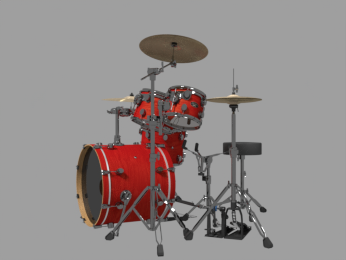}\\
    \includegraphics[height=2.15cm, clip=true]{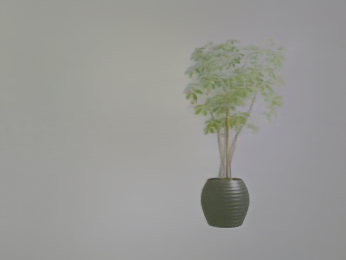}&
    \includegraphics[height=2.15cm, clip=true]{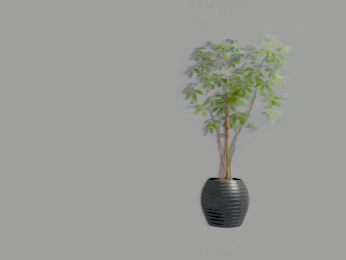}&
    \includegraphics[height=2.15cm, clip=true]{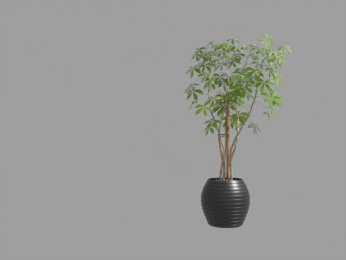}&
    \includegraphics[height=2.15cm, clip=true]{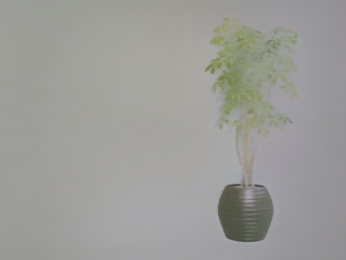}&
    \includegraphics[height=2.15cm, clip=true]{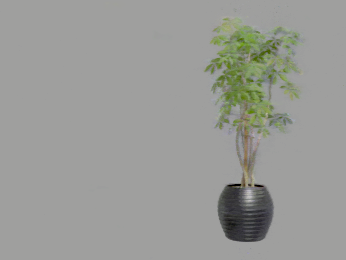}&
    \includegraphics[height=2.15cm, clip=true]{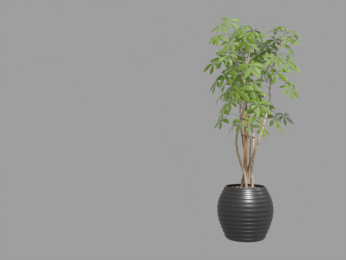}\\
    \includegraphics[height=2.15cm, clip=true]{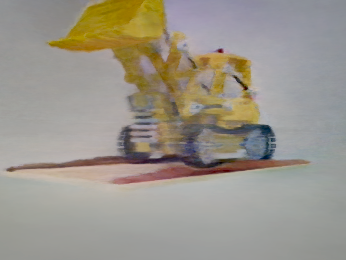}&
    \includegraphics[height=2.15cm, clip=true]{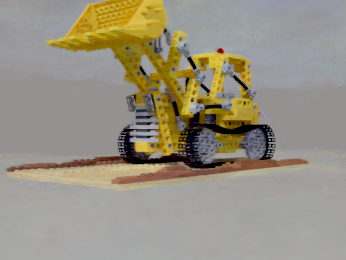}&
    \includegraphics[height=2.15cm, clip=true]{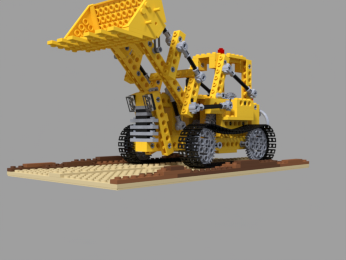}&
    \includegraphics[height=2.15cm, clip=true]{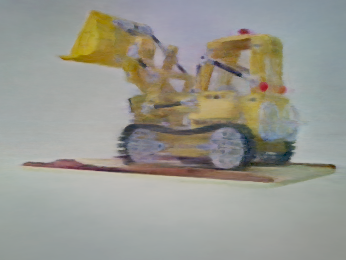}&
    \includegraphics[height=2.15cm, clip=true]{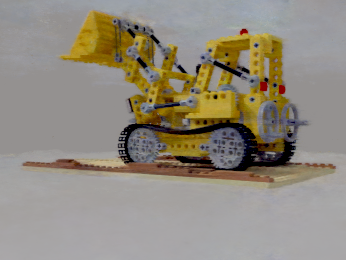}&
    \includegraphics[height=2.15cm, clip=true]{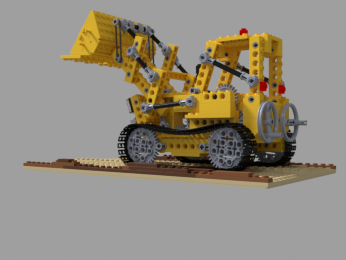}\\
    \includegraphics[height=2.15cm, clip=true]{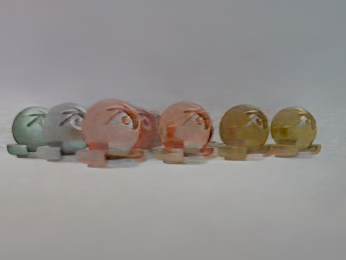}&
    \includegraphics[height=2.15cm, clip=true]{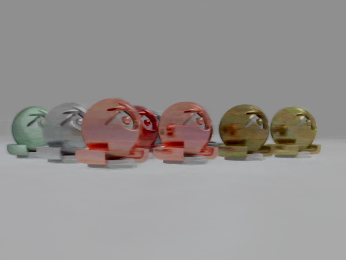}&
    \includegraphics[height=2.15cm, clip=true]{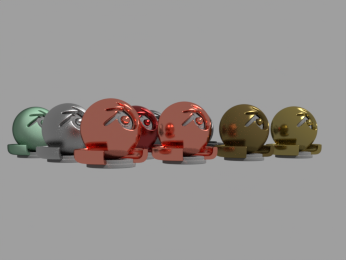}&
    \includegraphics[height=2.15cm, clip=true]{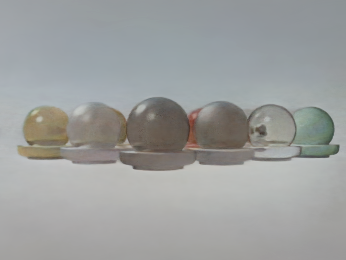}&
    \includegraphics[height=2.15cm, clip=true]{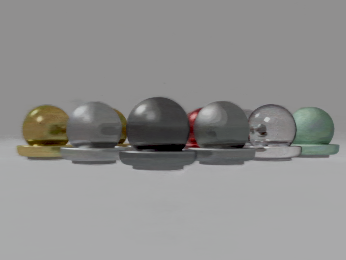}&
    \includegraphics[height=2.15cm, clip=true]{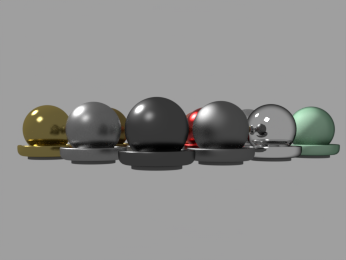}\\
    \includegraphics[height=2.15cm, clip=true]{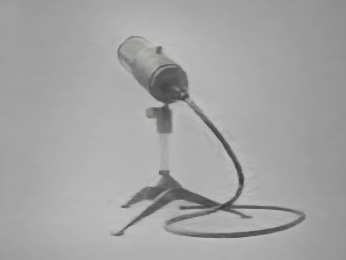}&
    \includegraphics[height=2.15cm, clip=true]{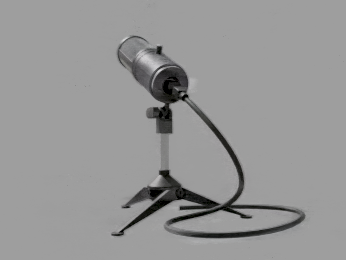}&
    \includegraphics[height=2.15cm, clip=true]{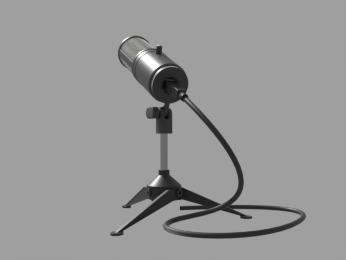}&
    \includegraphics[height=2.15cm, clip=true]{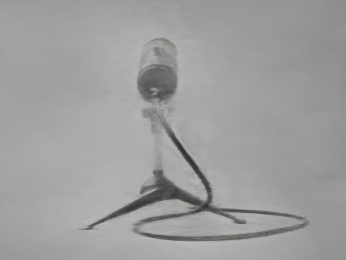}&
    \includegraphics[height=2.15cm, clip=true]{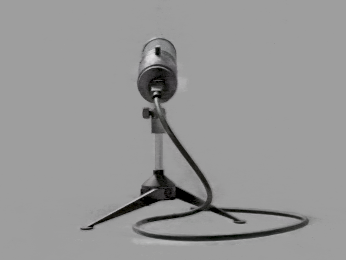}&
    \includegraphics[height=2.15cm, clip=true]{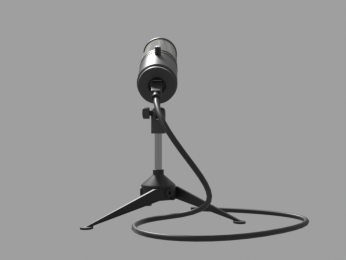}\\

       \small{E2VID+NeRF} & \small{EventNeRF} & \small{Ground Truth} & \small{E2VID+NeRF} & \small{EventNeRF} & \small{Ground Truth} \vspace{0.1cm}
    \end{tabular}

    \caption{Additional results and E2VID~\cite{Rebecq19pami}+NeRF~\cite{mildenhall2020nerf} comparisons on all of the used synthetic scenes (from top: Hotdog, Chair, Drums, Ficus, Lego, Materials and Microphone).
    }
    \label{fig:moresynthetic}
\end{figure*}%

\section{Real Data Results}
\label{sec:real}
We provide more real data results in Fig.~\ref{fig:morereal} and in the supplementary video.
In the ``Plant'' scene, we can reconstruct every stem and thin leaf.
In the ``Sewing'' scene, we recover even a one-pixel-wide needle of the machine (best viewed with zoom).
In the ``Microphone'' scene, we can reconstruct fine details such as a microphone grid. 
In the ``Controller'' scene, we preserve its details in the dark regions despite having low contrast.
Similarly, in the ``Goatling'' scene, we can reconstruct both the details in the dark and bright highlights on the glasses.
In the ``Cube'' scene, EventNeRF recovers sharp colour details.
``Multimeter'', ``Cube'' and ``Sewing'' show how we recover view-dependent effects.
In the ``Bottle'' scene, we can see the drawings on the reconstructed label.
All our results are halo-free.
\begin{figure*}[ht]
    \centering
    \begin{tabular}{@{}c@{\hspace{0.05cm}}c@{\hspace{0.05cm}}c@{\hspace{0.05cm}}c@{}}
    \includegraphics[height=3.2cm, clip=true]{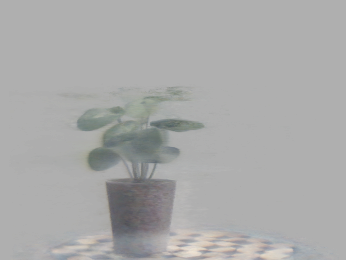}&
    \includegraphics[height=3.2cm, clip=true]{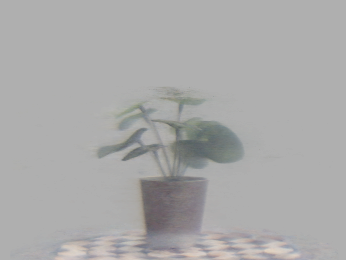}&
    \includegraphics[height=3.2cm, clip=true]{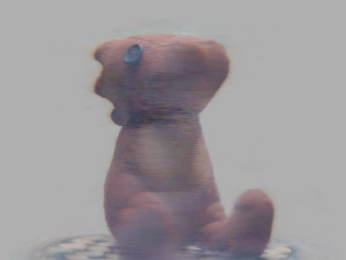}&
    \includegraphics[height=3.2cm, clip=true]{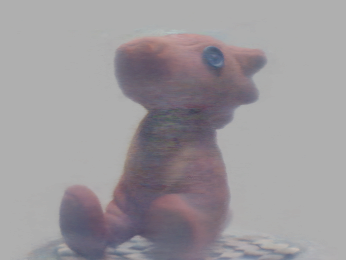}\\
    \multicolumn{2}{c}{Plant} & \multicolumn{2}{c}{Dragon}\\
    \includegraphics[height=3.2cm,    clip=true]{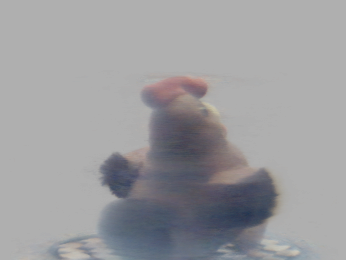}&
    \includegraphics[height=3.2cm, clip=true]{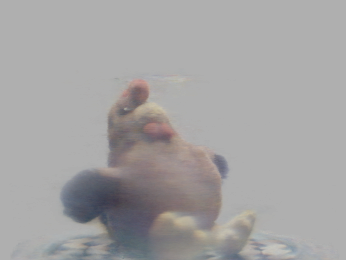}&
    \includegraphics[height=3.2cm, clip=true]{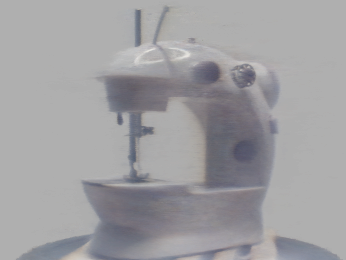}&
    \includegraphics[height=3.2cm, clip=true]{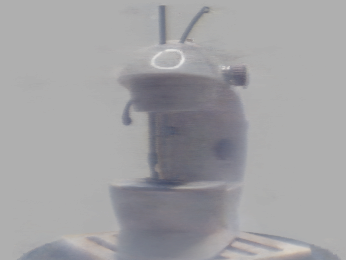}\\
    \multicolumn{2}{c}{Chicken} & \multicolumn{2}{c}{Sewing}\\
    \includegraphics[height=3.2cm,    clip=true]{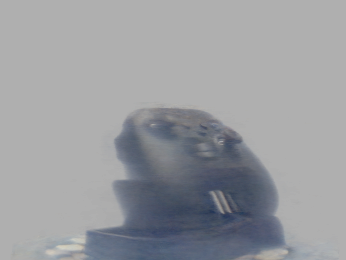}&
    \includegraphics[height=3.2cm, clip=true]{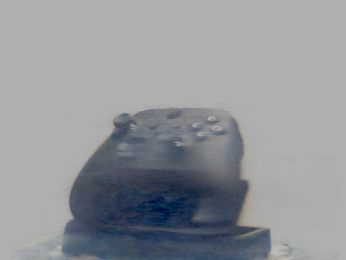}&
    \includegraphics[height=3.2cm, clip=true]{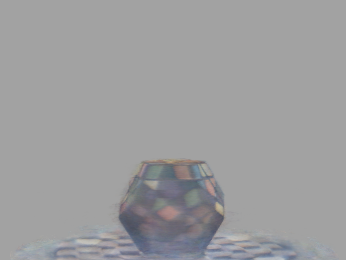}&
    \includegraphics[height=3.2cm, clip=true]{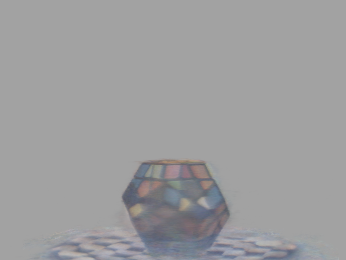}
    \\\multicolumn{2}{c}{Controller} & \multicolumn{2}{c}{Cube}\\
    \includegraphics[height=3.2cm,    clip=true]{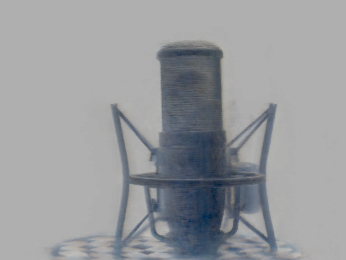}&
    \includegraphics[height=3.2cm, clip=true]{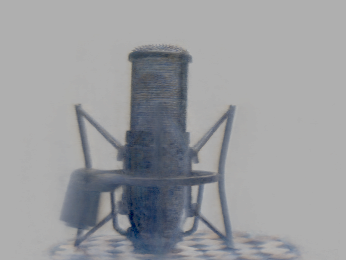}&
    \includegraphics[height=3.2cm, clip=true]{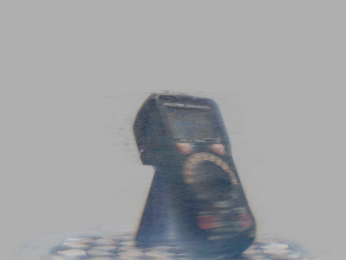}&
    \includegraphics[height=3.2cm, clip=true]{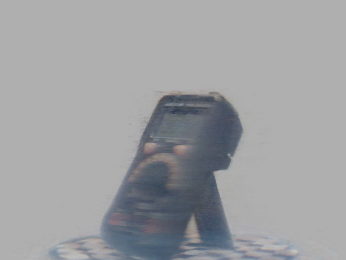}\\
    \multicolumn{2}{c}{Microphone} & \multicolumn{2}{c}{Multimeter}\\
    \includegraphics[height=3.2cm,    clip=true]{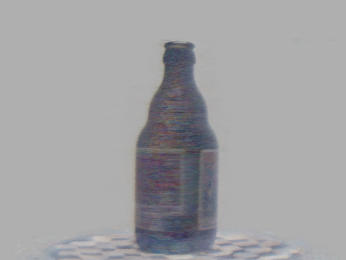}&
    \includegraphics[height=3.2cm, clip=true]{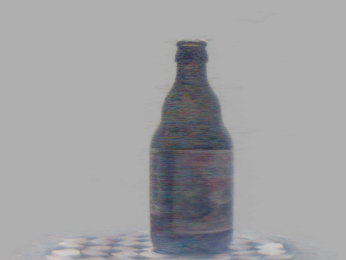}&
    \includegraphics[height=3.2cm, clip=true]{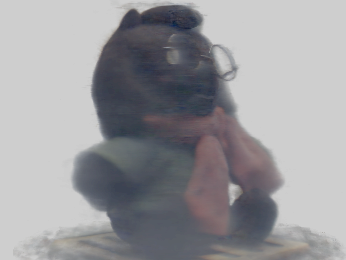}&
    \includegraphics[height=3.2cm, clip=true]{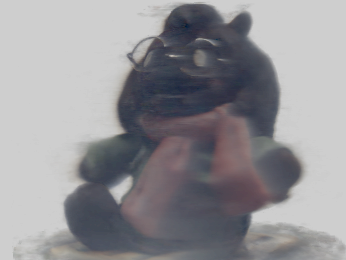}\\
    \multicolumn{2}{c}{Bottle} & \multicolumn{2}{c}{Goatling}
 \vspace{0.1cm}
    \end{tabular}
    \caption{Additional EventNeRF reconstructions of the real scenes. We show two arbitrary views per scene.
    }
    \label{fig:morereal}
\end{figure*}%

\begin{figure*}[ht]
    \centering
    \begin{tabular}{@{}c@{\hspace{0.05cm}}c@{\hspace{0.05cm}}c@{\hspace{0.05cm}}c@{}}
    \includegraphics[height=3.2cm, clip=true]{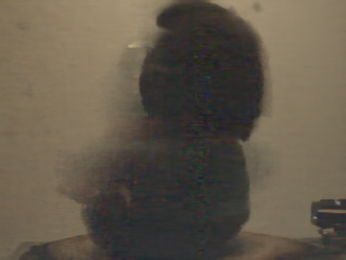}&
    \includegraphics[height=3.2cm, clip=true]{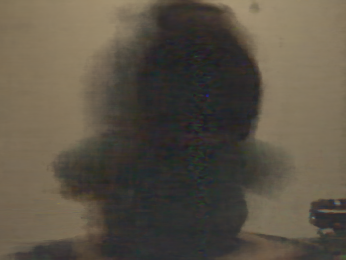}&
    \includegraphics[height=3.2cm, clip=true]{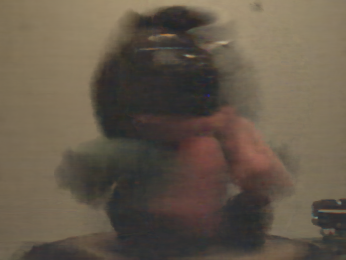}&
    \includegraphics[height=3.2cm, clip=true]{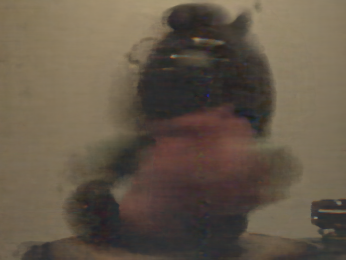}
    \\\multicolumn{4}{c}{Deblur-NeRF~\cite{ma2021deblur}} \\
    \includegraphics[height=3.2cm,    clip=true]{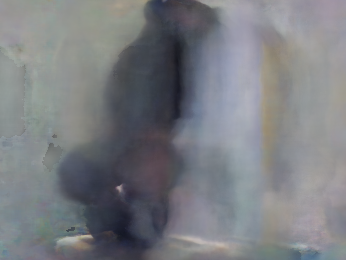}&
    \includegraphics[height=3.2cm, clip=true]{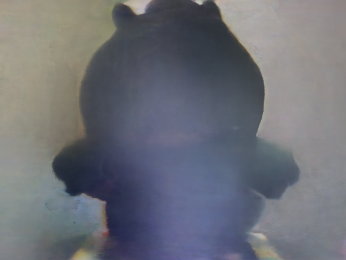}&
    \includegraphics[height=3.2cm, clip=true]{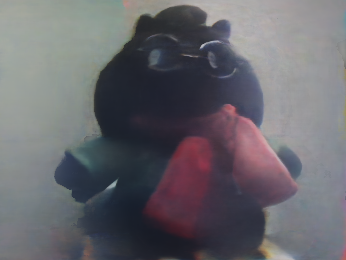}&
    \includegraphics[height=3.2cm, clip=true]{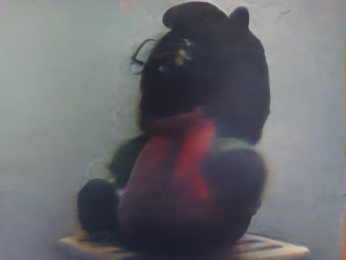}
    \\\multicolumn{4}{c}{E2VID~\cite{Rebecq19pami}+NeRF~\cite{mildenhall2020nerf}}\\
    \includegraphics[height=3.2cm,    clip=true]{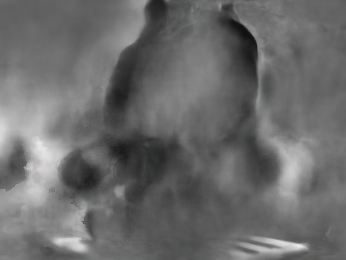}&
    \includegraphics[height=3.2cm, clip=true]{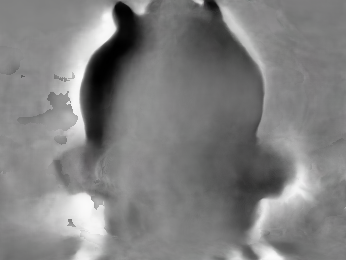}&
    \includegraphics[height=3.2cm, clip=true]{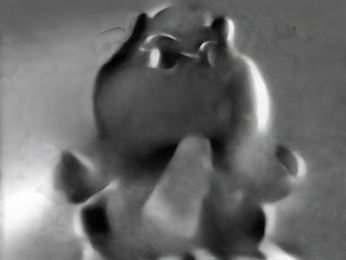}&
    \includegraphics[height=3.2cm, clip=true]{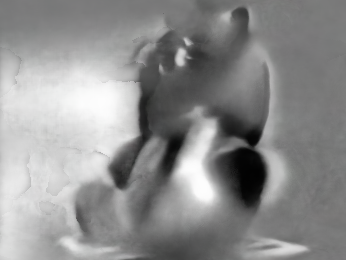}
    \\\multicolumn{4}{c}{ssl-E2VID~\cite{Paredes21}+NeRF~\cite{mildenhall2020nerf}}\\
    \includegraphics[height=3.2cm,    clip=true]{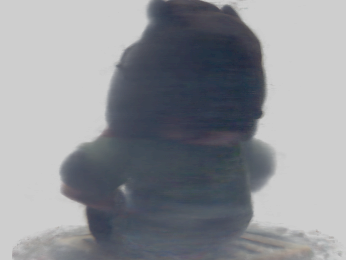}&
    \includegraphics[height=3.2cm, clip=true]{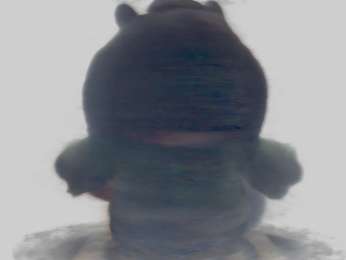}&
    \includegraphics[height=3.2cm, clip=true]{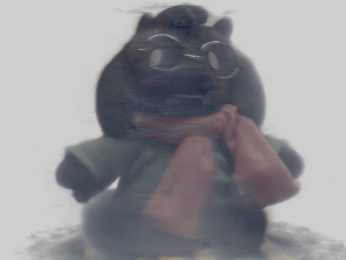}&
    \includegraphics[height=3.2cm, clip=true]{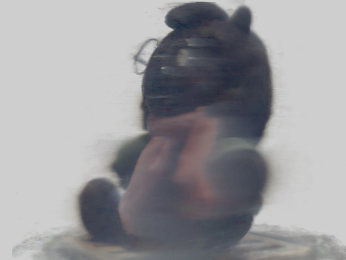}
    \\\multicolumn{4}{c}{{Our EventNeRF}}
 \vspace{0.1cm}
    \end{tabular}
    \caption{Results generated by different approaches. Our EventNeRF clearly outperforms all the methods. 
    }
    \label{fig:comparison1}
\end{figure*}

\begin{figure*}[ht]
    \centering
    \begin{tabular}{@{}c@{\hspace{0.05cm}}c@{\hspace{0.05cm}}c@{\hspace{0.05cm}}c@{}}
    \includegraphics[height=3.2cm, clip=true]{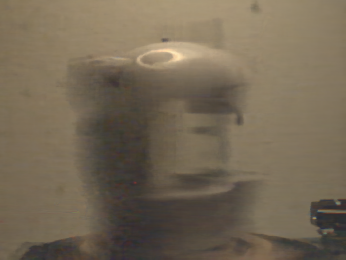}&
    \includegraphics[height=3.2cm, clip=true]{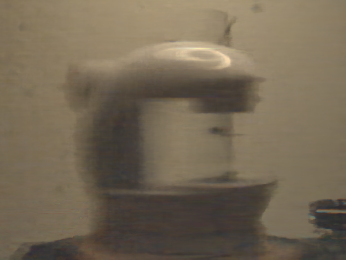}&
    \includegraphics[height=3.2cm, clip=true]{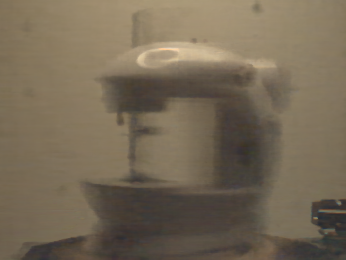}&
    \includegraphics[height=3.2cm, clip=true]{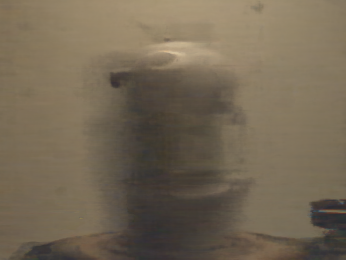}
    \\\multicolumn{4}{c}{Deblur-NeRF~\cite{ma2021deblur}} \\
    \includegraphics[height=3.2cm,    clip=true]{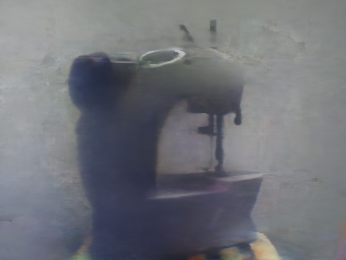}&
    \includegraphics[height=3.2cm, clip=true]{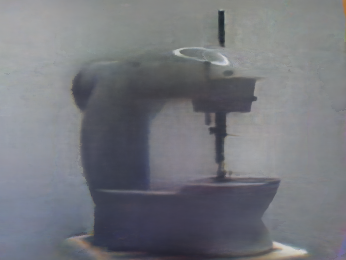}&
    \includegraphics[height=3.2cm, clip=true]{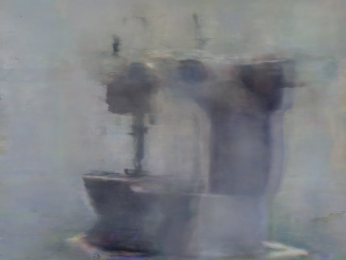}&
    \includegraphics[height=3.2cm, clip=true]{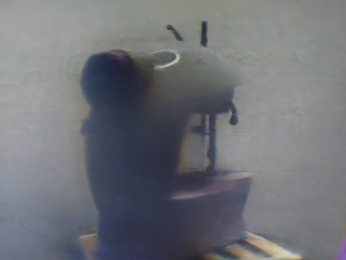}
    \\\multicolumn{4}{c}{E2VID~\cite{Rebecq19pami}+NeRF~\cite{mildenhall2020nerf}} \\
    \includegraphics[height=3.2cm,    clip=true]{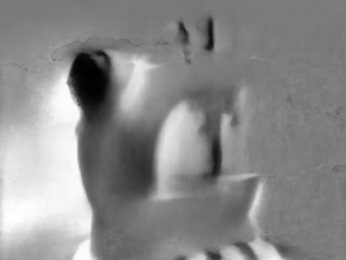}&
    \includegraphics[height=3.2cm, clip=true]{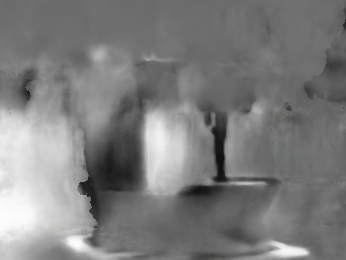}&
    \includegraphics[height=3.2cm, clip=true]{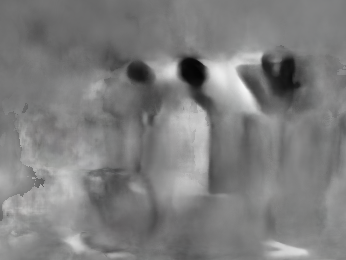}&
    \includegraphics[height=3.2cm, clip=true]{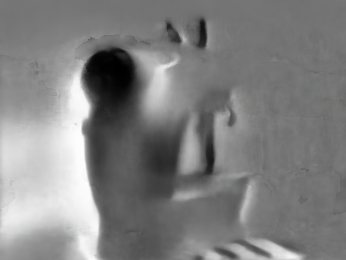}
    \\\multicolumn{4}{c}{ssl-E2VID~\cite{Paredes21}+NeRF~\cite{mildenhall2020nerf}} \\
    \includegraphics[height=3.2cm,    clip=true]{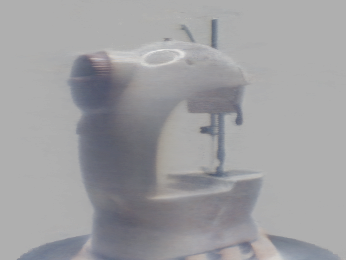}&
    \includegraphics[height=3.2cm, clip=true]{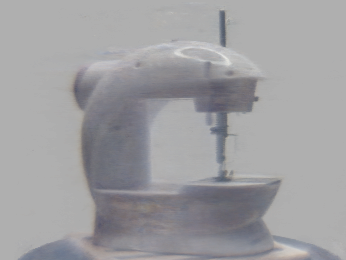}&
    \includegraphics[height=3.2cm, clip=true]{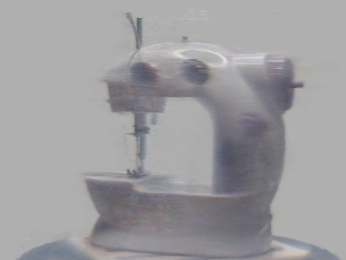}&
    \includegraphics[height=3.2cm, clip=true]{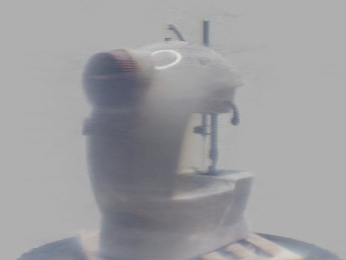}
    \\\multicolumn{4}{c}{{Our EventNeRF}}
 \vspace{0.1cm}
    \end{tabular}
    \caption{Results generated by different approaches. Our EventNeRF clearly outperforms all the methods. 
    }
    \label{fig:comparison2}
\end{figure*}

We show more visual results for DeblurNeRF~\cite{ma2021deblur} in Fig.~\ref{fig:comparison1} and Fig.~\ref{fig:comparison2} (top). As stated in the main text, Deblur-NeRF can not handle view-consistent blur and thus produces clearly worse results than our method.
We also show in the same figures visual results for E2VID~\cite{Rebecq19pami}+NeRF~\cite{mildenhall2020nerf} and ssl-E2VID~\cite{Paredes21}+NeRF~\cite{mildenhall2020nerf}.
As stated in the main text, results produced by ssl-E2VID are clearly worse than E2VID,
In addition, ssl-E2VID can only generate grayscale images.
Our approach, however, outperforms all related methods.

\begin{table}
\scriptsize
    \centering
\resizebox{\columnwidth}{!}{%
\begin{tabular}{c|c|c|c|c|c|c}
 & \multicolumn{3}{c}{Our EventNeRF} & \multicolumn{3}{c}{Our Real-time Implementation}\\
Scene & PSNR $\uparrow$ & SSIM $\uparrow$ & LPIPS $\downarrow$ & PSNR $\uparrow$ & SSIM $\uparrow$ & LPIPS $\downarrow$ \\
\hline
Drums & $\mathbf{27.43}$ & $\mathbf{0.91}$ & $\mathbf{0.07}$ & $26.03$ & $\mathbf{0.91}$ & $\mathbf{0.07}$\\
Lego & $\mathbf{25.84}$ & $\mathbf{0.89}$ & $0.13$ & $22.82$ & $\mathbf{0.89}$ & $\mathbf{0.08}$\\
Chair & $\mathbf{30.62}$ & $\mathbf{0.94}$ & $\mathbf{0.05}$ & $27.97$ & $\mathbf{0.94}$ & $\mathbf{0.05}$\\
Ficus & $\mathbf{31.94}$ & $\mathbf{0.94}$ & $\mathbf{0.05}$ & $26.77$ & $0.92$ & $0.12$\\
Mic & $\mathbf{31.78}$ & $\mathbf{0.96}$ & $\mathbf{0.03}$ & $28.34$ & $0.95$ & $0.04$\\
Hotdog & $\mathbf{30.26}$ & $\mathbf{0.94}$ & $\mathbf{0.04}$ & $23.99$ & $0.93$ & $0.10$\\
Materials & $24.10$ & $\mathbf{0.94}$ & $\mathbf{0.07}$ & $\mathbf{26.05}$ & $0.93$ & $\mathbf{0.07}$\\
\hline
Average & $\mathbf{28.85}$ & $\mathbf{0.93}$ & $\mathbf{0.06}$ & $25.99$ & $0.92$ & $0.07$
\end{tabular}
}
\caption{Comparing our method using the original NeRF implementation~\cite{mildenhall2020nerf} (EventNeRF) against a real-time implementation based on torch-ngp~\cite{torch-ngp}. While the real-time implementation takes significantly less training and testing time, it can compromise some of the rendering quality.}
\label{tbl:ngp}
\end{table}

\begin{figure*}[ht]
    \centering
    \begin{tabular}{@{}c@{\hspace{0.05cm}}c@{\hspace{0.05cm}}c@{\hspace{0.05cm}}c@{\hspace{0.05cm}}c@{\hspace{0.05cm}}c@{}}
    \includegraphics[height=2.15cm, clip=true]{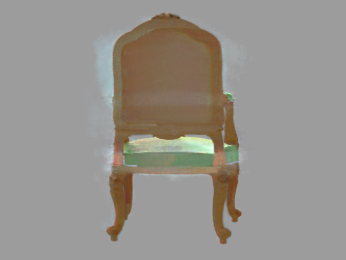}&
    \includegraphics[height=2.15cm, clip=true]{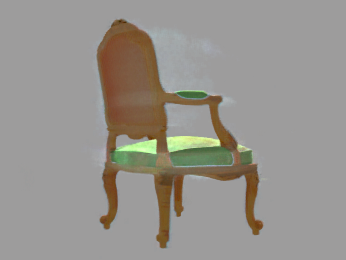}&
    \includegraphics[height=2.15cm, clip=true]{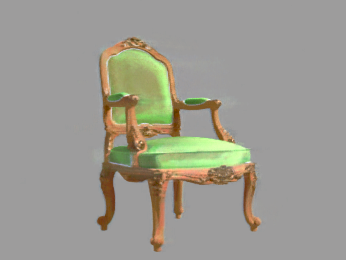}&
    \includegraphics[height=2.15cm, clip=true]{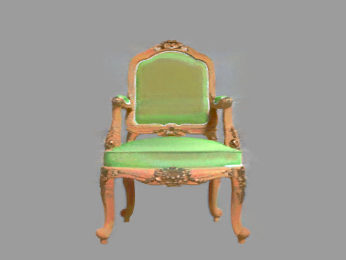}&
    \includegraphics[height=2.15cm, clip=true]{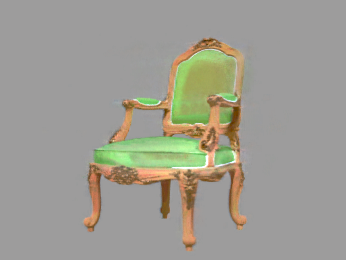}&
    \includegraphics[height=2.15cm, clip=true]{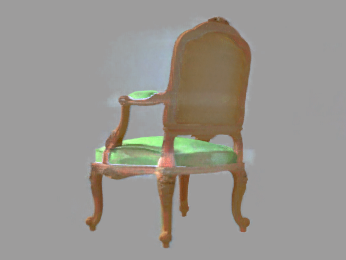}\\
    \includegraphics[height=2.15cm, clip=true]{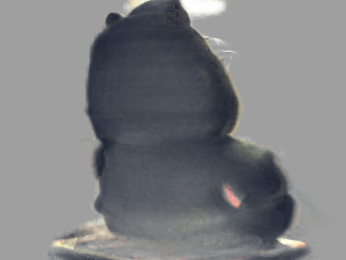}&
    \includegraphics[height=2.15cm, clip=true]{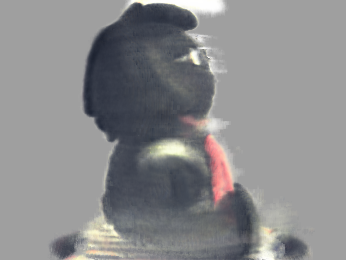}&
    \includegraphics[height=2.15cm, clip=true]{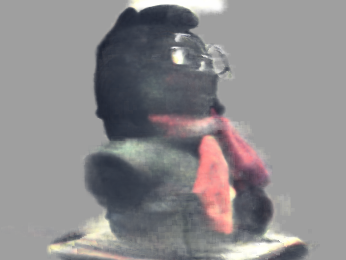}&
    \includegraphics[height=2.15cm, clip=true]{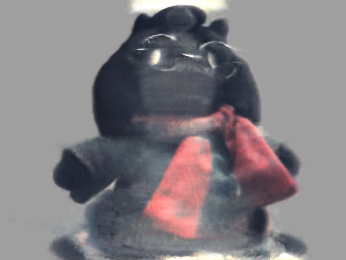}&
    \includegraphics[height=2.15cm, clip=true]{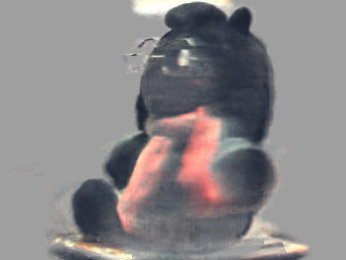}&
    \includegraphics[height=2.15cm, clip=true]{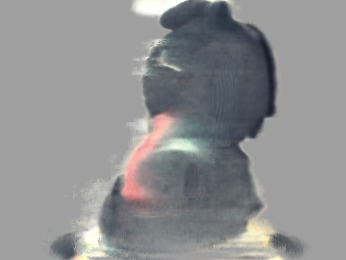}\\
    \includegraphics[height=2.15cm, clip=true]{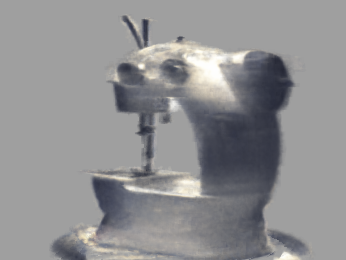}&
    \includegraphics[height=2.15cm, clip=true]{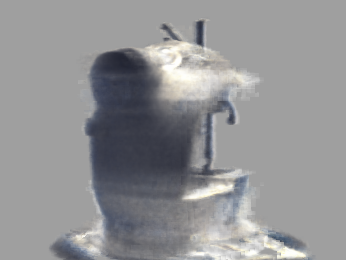}&
    \includegraphics[height=2.15cm, clip=true]{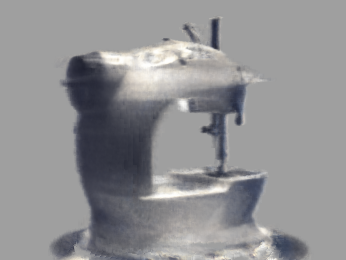}&
    \includegraphics[height=2.15cm, clip=true]{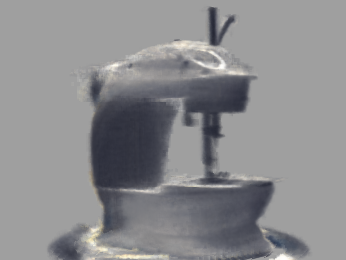}&
    \includegraphics[height=2.15cm, clip=true]{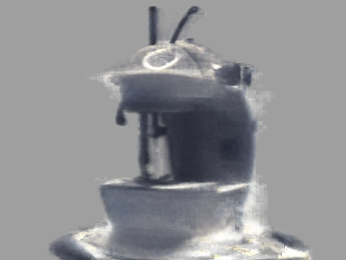}&
    \includegraphics[height=2.15cm, clip=true]{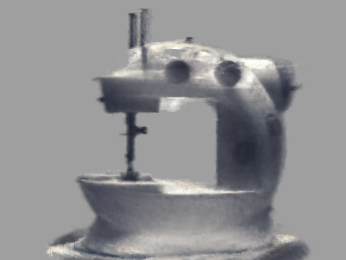}\\
    \end{tabular}
    \caption{Results on synthetic (top) and real sequences using our real-time implementation. This approach takes around a minute to train and runs in real-time during the test.
    }
    \label{fig:ngp}
\end{figure*}%

\begin{figure*}[ht]
    \centering
    \begin{tabular}{@{}c@{\hspace{0.05cm}}c@{\hspace{0.05cm}}c@{\hspace{0.05cm}}c@{}}
    \includegraphics[height=3.2cm, clip=true]{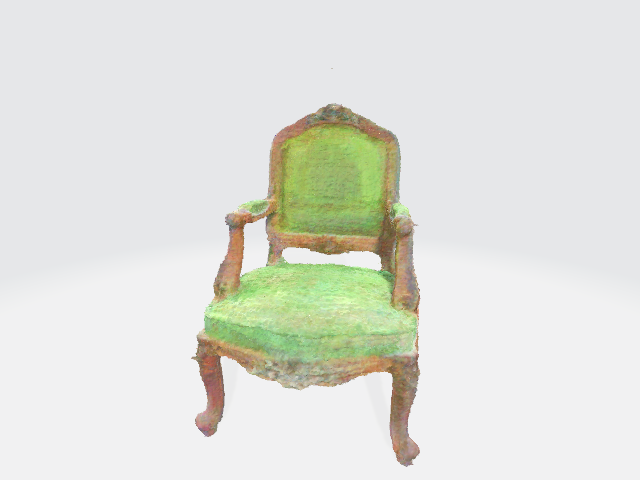}&
    \includegraphics[height=3.2cm, clip=true]{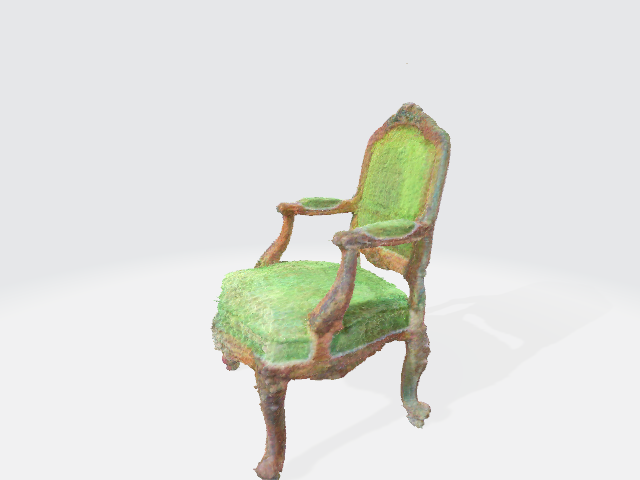}&
    \includegraphics[height=3.2cm, clip=true]{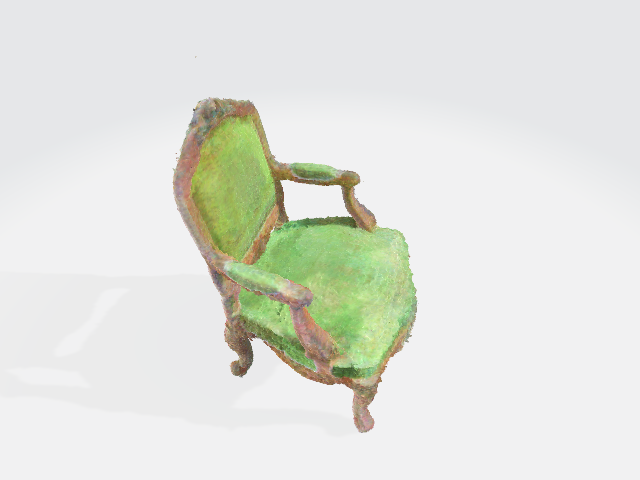}&
    \includegraphics[height=3.2cm, clip=true]{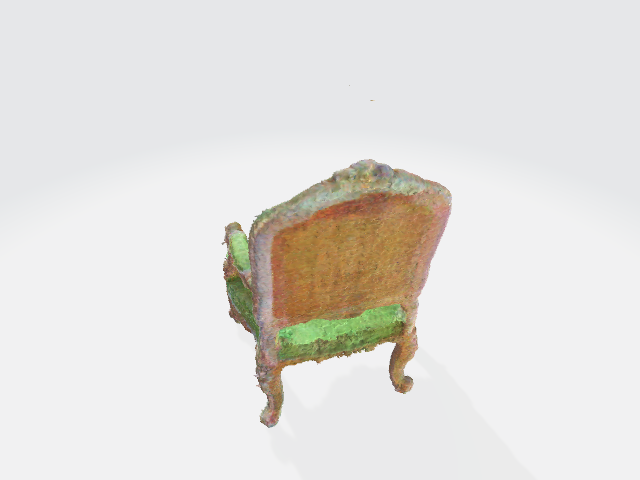}\\
    \end{tabular}
    \caption{We extract a textured mesh using marching cubes \cite{marchingcubes}. The mesh can be viewed from an arbitrary viewpoint. 
    }
    \label{fig:mesh}
\end{figure*}%

\section{Real-Time Implementation}
\label{sec:realtime}

We show an interactive application of our approach that runs in real-time. For this, we implement our method using torch-ngp~\cite{torch-ngp} instead of the original NeRF representation~\cite{mildenhall2020nerf}.
Training a model using this implementation takes around a minute using a single NVIDIA GeForce RTX 2070 GPU.
At the test time, the method runs in real-time.
We show in Fig.~\ref{fig:ngp} visual results for this implementation.
For image sequence results, please refer to the supplemental video.
Our real-time implementation produces highly photorealistic results that can be viewed from an arbitrary viewpoint.
This, however, can come with some trade-off in the rendering quality as shown in Tab.~\ref{tbl:ngp}.

\section{Tone-Mapping in Different Scenarios} 
The primary difference between the appearance of real, synthetic and the real-time results is the tone-mapping used.
As real events have no ground-truth tone-mapping, we tune it manually.
To show the details in the dark regions we opt for the reduced contrast in Fig.~\ref{fig:SynReal} which results in images that could look washed out. 
The speed gain of the real-time implementation is from using torch-ngp~\cite{torch-ngp}, which is significantly faster than the original NeRF~\cite{mildenhall2020nerf}.

\section{Mesh Extraction}
\label{sec:mesh}

As another application, we show in Fig.~\ref{fig:mesh} that we can 
extract a mesh from our NeRF reconstruction using marching  cubes~\cite{marchingcubes}. 
The extracted mesh is textured and can be rendered from an arbitrary camera viewpoint.
For more results, please see the interactive demo in our video.

\section{Event Window Temporal Bounds Sampling} 
\label{sec:win_sampling} 
In Sec.~\ref{sec:ray_sampling}, the ends of the time intervals $t$ of all windows are fixed and uniformly distributed through the whole length of the stream.
This way, all views are sampled uniformly.
As our recordings are perfect loops, we concatenate corresponding events from the end when there are not enough events from the start of the stream for the sampled window length.
Window lengths $L_\mathrm{min}$ and $L_\mathrm{max}$ were chosen empirically as the highest and lowest values that did not result in the model diverging.

\begin{figure*}[ht]
    \centering

\begin{tabular}{@{}c@{\hspace{0.05cm}}c@{\hspace{0.05cm}}c@{\hspace{0.05cm}}c@{}}
    \includegraphics[height=3.2cm, clip=true]{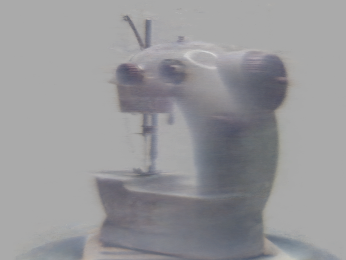}&
    \includegraphics[height=3.2cm, clip=true]{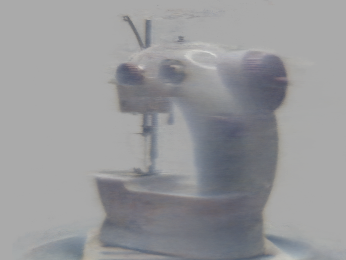}&
    \includegraphics[height=3.2cm, clip=true]{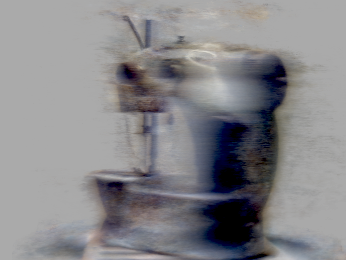}&
    \includegraphics[height=3.2cm, clip=true]{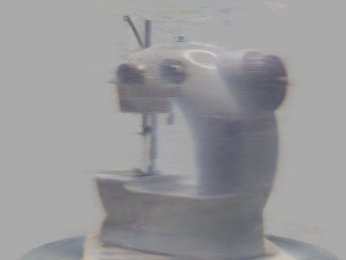}\\

    \includegraphics[height=3.2cm, clip=true]{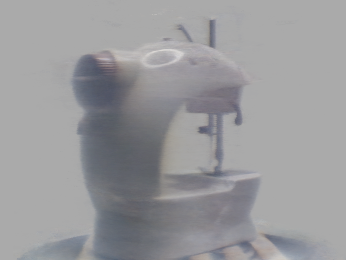}&
    \includegraphics[height=3.2cm, clip=true]{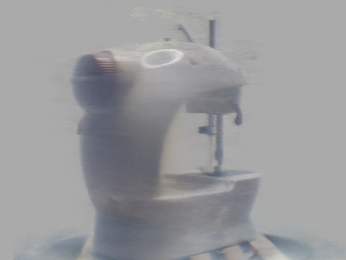}&
    \includegraphics[height=3.2cm, clip=true]{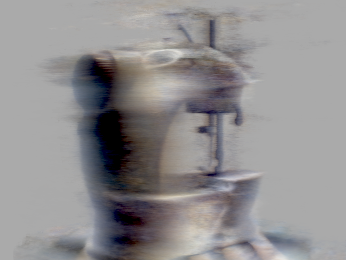}&
    \includegraphics[height=3.2cm, clip=true]{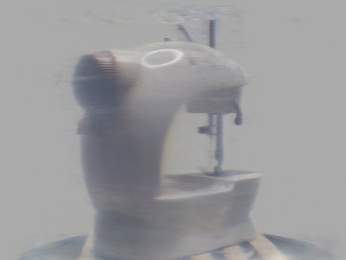}\\

       \small{Full EventNeRF} & \small{No negative sampling} & \small{Const.~short 10ms windows} & \small{Const.~long 100ms windows}
       \vspace{0.1cm}
    \end{tabular}

    \caption{Ablation studies on real data for the ``Sewing'' scene. The full model provides the best results with the most detail and fewest artefacts. In case of ``No negative sampling'', note the artefacts above and around the object. In the case of constant long windows, note the blurriness. For the detailed analysis of these results, please refer to Sec.~\ref{sec:ablation_real}.
    }
    \label{fig:ablation_real}
\end{figure*}%

\section{Ablation Study on Real Data}
\label{sec:ablation_real}

We ran the same ablation studies as in Sec.~\ref{sec:ablation} on real data; see Fig.~\ref{fig:ablation_real} for the qualitative results on the ``Sewing'' sequence. 
Our full method produces the best results.
Using short constant window length instead of a randomised one results in significant artefacts.
As only views close to each other are used to supervise the model in this case, long-term consistency and low-frequency lighting does not propagate well.
Using long constant window length leads to the noticeable blur in the reconstruction as short-time details and high-frequency lighting information is not present in long windows.
Using only positive sampling causes more artefacts than in the full model with negative sampling. 
We note that this effect is less severe than in the case of synthetic data in Fig.~\ref{fig:E2VID-real}. 
The differences are perhaps explained by the inaccuracies in the camera parameters for the real data. 
If the camera poses are highly accurate, the negative sampling becomes more important; otherwise, the differences will be smaller than the artefacts caused by the inaccuracies in camera parameters. 
Nevertheless, these differences still can be noticed in both real and synthetic data models.  
To explore this hypothesis further, we analyse the effect of introducing error to the camera parameters in the next section.

\section{Robustness to Camera Pose Errors}
\label{sec:abl_angle}
\begin{figure*}[ht]
    \centering

\begin{tabular}{@{}c@{\hspace{0.05cm}}c@{\hspace{0.05cm}}c@{\hspace{0.05cm}}c@{\hspace{0.05cm}}c@{}}
    \includegraphics[height=2.55cm, clip=true]{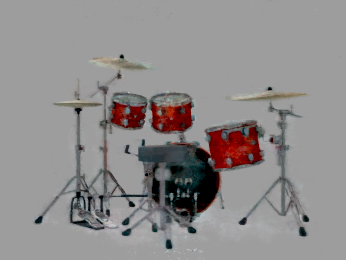}&
    \includegraphics[height=2.55cm, clip=true]{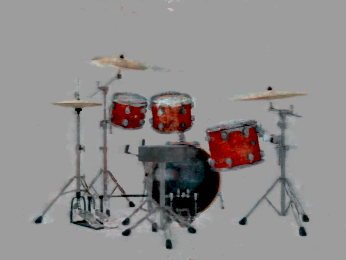}&
    \includegraphics[height=2.55cm, clip=true]{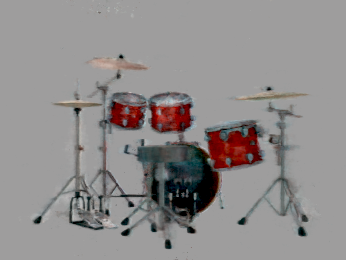}&
    \includegraphics[height=2.55cm, clip=true]{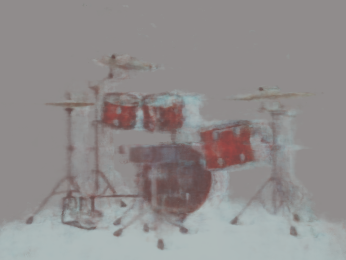}&
    \includegraphics[height=2.55cm, clip=true]{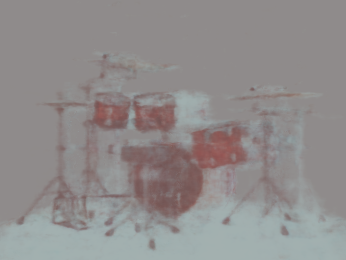}\\

       \small{$0^\circ$ ($27.43$ dB PSNR)} & \small{$0.01^\circ$ ($27.26$ dB PSNR)} & \small{$0.1^\circ$ ($26.18$ dB PSNR)} & \small{$1^\circ$ ($18.11$ dB PSNR)} & \small{$2^\circ$ ($17.49$ dB PSNR)}
       \vspace{0.1cm}
    \end{tabular}
    \caption{Study on the robustness of EventNeRF to the added error in camera parameters on ``Drums''. Rotation error (specified in degrees) is added to the camera extrinsics to simulate inaccuracies in camera pose calibration described in Sec.~\ref{sec:cpc}. The results suggest that above a certain threshold between $0.1^\circ$ and $1^\circ$, the reconstruction quality starts to rapidly decay due to the trailing artefacts. Please refer to Sec.~\ref{sec:abl_angle} for our detailed analysis of these results.
    }
    \label{fig:camera_pose}
\end{figure*}%
\begin{figure*}[ht!]
    \centering

\begin{tabular}{@{}c@{\hspace{0.05cm}}c@{\hspace{0.05cm}}c@{\hspace{0.05cm}}c@{}}
    \includegraphics[height=3.2cm, clip=true]{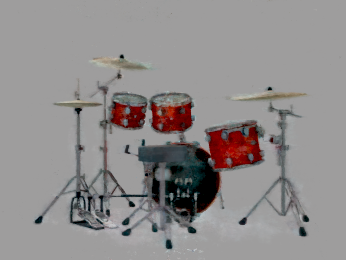}&
    \includegraphics[height=3.2cm, clip=true]{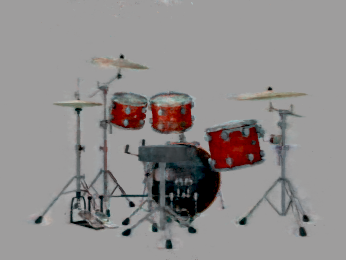}&
    \includegraphics[height=3.2cm, clip=true]{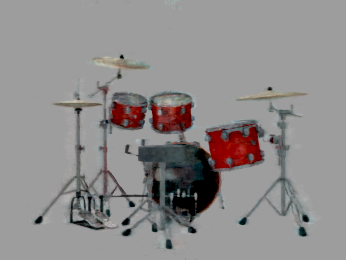}&
    \includegraphics[height=3.2cm, clip=true]{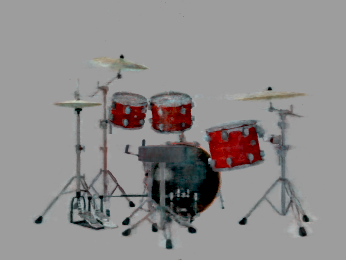}\\

       \small{$0\%$ ($27.43$ dB PSNR)} & \small{$1\%$ ($27.35$ dB PSNR)} & \small{$5\%$ ($27.36$ dB PSNR)} & \small{$15\%$ ($27.31$ dB PSNR)}
       \vspace{0.1cm}
    \end{tabular}

    \caption{Study on the robustness of EventNeRF to the added noise events on ``Drums''. A uniform number of random events (specified in percentages) is added to the training event stream to simulate noise found in real data. The subtlety of the differences suggests that our model is robust to noise events in amounts found within real data captures. Please refer to Sec.~\ref{sec:abl_noise} for our detailed analysis of these results. 
    } 
    \label{fig:event_noise} 
\end{figure*} 

We noticed that camera pose errors lead to trailing artefacts. 
Hence, we developed a camera pose calibration technique for our setup  which we describe in detail in Sec.~\ref{sec:cpc}. 
This led to cleaner predictions. 

To measure this effect, we introduce error into synthetic data camera poses and measure the performance on ``Drums'' (see Fig.~\ref{fig:camera_pose}). 
$0^\circ$, $0.01^\circ$, $0.1^\circ$, $1^\circ$, $2^\circ$ errors results in 27.43, 27.26, 26.18, 18.11, 17.49 dB PSNR, correspondingly ($1^\circ$ translates to 10-15 pixels offset at 346$\times$260 pixels image resolution).
The steep change starting at $1^\circ$ is due to significant increase in trailing artefacts. 
This suggests that at some threshold level of camera pose error between $0.1^\circ$ and $1^\circ$, the reconstruction quality starts to rapidly degrade.

\section{Robustness to Noisy Events}
\label{sec:abl_noise}

For real data, we measure the amount of noise when recording a static scene with a static camera.
We record the number of noise events per second ($\mathrm{ev}/\mathrm{s}$) using the lowest and the highest brightness settings of our light source.
Both settings result in similar noise measurements of around $1.1\cdot 10^5 ~\mathrm{ev}/\mathrm{s}$.
Hence, we believe the amount of noise with our camera settings is almost constant regardless of the scene brightness.
In our real data experiments, this amounts to $10-18\%$ of the training event streams being noise.

With improved lighting, we do not expect our results to change  significantly. 
The event camera is reporting in the log-space and, hence, would  emit the same number of events as long as the contrast between the darkest and brightest parts is the same. 

For synthetic data, we add no noise to the event stream in the main text. 
However, if we add $1$, $5$, $15\%$ of noisy events to the training event stream, we obtain 27.35, 27.36, 27.31 dB PSNR on ``Drums'' (w/o noise: 27.43 dB). 
The visualisations of the experimental results given in Fig.~\ref{fig:event_noise} also confirm that the change in quality is slight between all tested models. 
This suggests that our method is sufficiently robust to event noise in the training data and that it is not the primary cause of artefacts in the case of real data.

\end{document}